\documentclass[10pt,journal,compsoc]{IEEEtran}

\usepackage{graphicx}
\usepackage{ifsym}
\usepackage{amssymb}
\usepackage{booktabs}
\usepackage{longtable}
\usepackage{makecell}
\usepackage{multirow}
\usepackage{flushend}
\usepackage{transparent}

\usepackage[dvipsnames]{xcolor}
\usepackage{bbding}
\usepackage{colortbl}
\usepackage{indentfirst}
\usepackage{subfig}

\usepackage[linesnumbered,ruled]{algorithm2e}
\usepackage{tabularx}
\usepackage{pifont}

\usepackage[export]{adjustbox}
\ifCLASSOPTIONcompsoc
  \usepackage[nocompress]{cite}
\else
  \usepackage{cite}
\fi

\ifCLASSINFOpdf

\else
  
\fi
\usepackage{amsmath}

\usepackage{hyperref}
\begin{document}
%
\title{Paving the Way for Point Cloud Video Representation Learning Using A PDE Model}
 
%
%
%
%

\author{Zhuoxu~Huang,
        Zhenkun~Fan,
        Jungong~Han\textsuperscript{\Envelope},~\IEEEmembership{Senior Member,~IEEE,}
        and Josef~Kittler,~\IEEEmembership{Life Member,~IEEE}
%
\IEEEcompsocitemizethanks{
\IEEEcompsocthanksitem This work was supported in part by National Natural Science Foundation of China No. 62441235 and No. 92570204, and is also supported by Beijing Natural Science Foundation (L257005).
\IEEEcompsocthanksitem Zhuoxu Huang and Zhenkun Fan are with the Department of Computer Science, Aberystwyth University, Aberystwyth SY23 3DB, U.K. (e-mail: zhh6@aber.ac.uk; zhf1@aber.ac.uk).
\IEEEcompsocthanksitem Jungong Han is with the Department of Automation, and is also with Beijing National Research Center for Information Science and Technology, Tsinghua University, Beijing, 100084, China. (e-mail: jungonghan77@gmail.com).
\IEEEcompsocthanksitem Josef Kittler is with the Department of Electrical Engineering, Surrey University, Surrey GU2 7XH, U.K. (e-mail: j.kittler@surrey.ac.uk).
\IEEEcompsocthanksitem \textsuperscript{\Envelope} Corresponding author: Jungong Han
}
}

%
%

\markboth{IEEE TRANSACTIONS ON PATTERN ANALYSIS AND MACHINE INTELLIGENCE}%
{Shell \MakeLowercase{\textit{et al.}}: Bare Demo of IEEEtran.cls for Computer Society Journals}
%

\IEEEtitleabstractindextext{%
\begin{abstract}
Investigating spatial-temporal correlations, specifically how spatial points vary over time, is crucial for understanding point cloud videos. Traditional methods, particularly flow-based techniques, struggle with these correlations due to the unordered spatial arrangement of sequential point cloud data. To address this challenge, we propose a novel approach that regularizes spatial-temporal correlation learning by formulating the problem as a solvable Partial Differential Equation (PDE). While PDEs have long been effective in the physical domain, their application to novel sequential data like point cloud video remains underexplored. Inspired by fluid analysis, we construct a simplified PDE, and the process of solving PDE is guided and refined by a contrastive learning structure between the temporal embeddings and the spatial embeddings. 
With this extra supervision, our method, named \textbf{MotionPDE}, serves as an effective, plug-and-play enhancement module for existing backbone models, adding minimal computational overhead and parameters.
Capitalizing on the contrastive learning process, we delve deeper into the self-supervised capabilities of MotionPDE, yielding promising results that underscore its utility and adaptability in point cloud video data interpretation. The code repo with trained checkpoints will be available \href{https://github.com/zhh6425/motionpde.git}{https://github.com/zhh6425/motionpde.git} for facilitating future research. 
\end{abstract}

}

\maketitle

\IEEEdisplaynontitleabstractindextext
\IEEEpeerreviewmaketitle

\section{Introduction}

\IEEEPARstart{S}{equential} data containing temporal dynamics provides a crucial context for understanding variations in our world. To model the inherent spatial-temporal correlations in such sequential data, partial differential equations (PDE) are widely used as a powerful tool in various domains including weather forecasting \cite{pdeStan, verma2024climode}, solid and fluid analyses \cite{fanaskov2023spectral, wu2023LSM, liu2023htnet}, image processing \cite{pdevision, mang2018pde, zhang2023pdeevent, Boquet2023pdeopticalflow}, and also video prediction \cite{guen2020disentangling, wu2023predictive}. 

\begin{figure}[htbp]
\begin{center}
\includegraphics[width=0.9\linewidth]{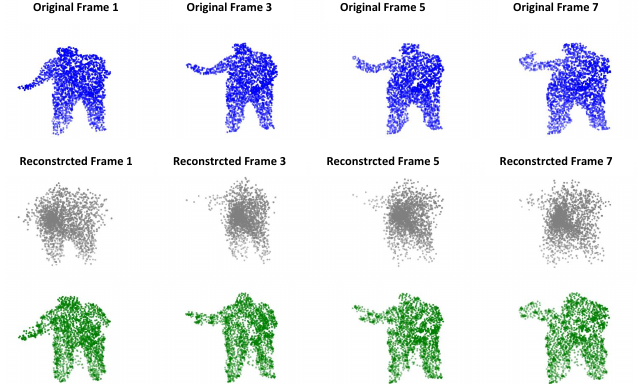}
\end{center}
\vspace{-0.3cm}
\caption{Motivation 1: Extra supervision related to spatial-temporal correlations enhances feature quality, leading to better reconstruction results and improved model performance. The first row is the \textcolor{blue}{ground truth}, followed by the reconstruction result of the \textcolor{gray}{baseline} \cite{fan2021pstnet} and the \textcolor[rgb]{0., 0.5, 0.}{pre-trained model with MotionPDE} (from Section \ref{sec:pretrain}).}
\label{fig:reconstruction_0}
\vspace{-0.5cm}
\end{figure}

Point cloud video, a cutting-edge sequential data format, enables accurate geometric perception with robust privacy protection. Consequently, it has become a primary tool in fields such as autonomous driving and robotics. 3D point clouds can be obtained from LiDAR, Radar, or RGB+D devices. They consist of a set of point coordinates $\{(x,y,z)\}$ that represent the geometric information of the objects. While these geometric points are arranged unordered spatially, a series of point cloud frames represents an ordered variation in temporal space. This unique format brings challenges to the spatial-temporal correlation learning of point cloud video, which is at the forefront of point cloud video modeling, similar to other sequential data. 

We argue that incorporating additional supervisory information related to spatial-temporal correlations can enhance the quality of features that are instrumental to point cloud video understanding. Therefore, it is logical to explore tools like PDE to improve point cloud video representation learning. However, to our knowledge, the integration of the PDE model into the point cloud video framework remains underdeveloped. 
To fill this gap, we propose \textit{MotionPDE, an enhancement model, aiming to regularize the spatial-temporal correlations learning for existing methods.}
\IEEEpubidadjcol

To solidify our motivation, we address two key points: (1) the problem of incorporating spatial-temporal correlations as an essential source of information for understanding point cloud videos, and (2) the rationale for choosing PDE over other tools for regularization, as well as the incompatibility of existing PDE models with point cloud videos. 

For the first point, we use qualitative visualizations to illustrate the abstract concept of spatial-temporal correlations. In the context of point cloud video data, reconstruction quality often serves as a practical proxy for evaluating the effectiveness of learned representations, as it reflects the model’s ability to preserve essential spatial-temporal structures during encoding. This perspective is supported by prior studies in autoencoder-based and self-supervised representation learning \cite{achlioptas2018learning, yang2018foldingnet, srivastava2015unsupervised}, which consistently demonstrate that better reconstruction generally correlates with more informative features. By freezing the trained model and adopting a reconstruction head on top of the backbone, we visualize the reconstruction results of the baseline and our MotionPDE-trained model in Figure \ref{fig:reconstruction_0} (full details can be found in Section \ref{sec:reconstructionresult}). The baseline model \cite{fan2021pstnet} fails to reconstruct the original spatial information, resulting in only a blurred structure (depicted in gray). However, with the additional spatial-temporal correlations provided by our MotionPDE, the model is capable of reconstructing detailed actions, such as waving hands (shown in green), demonstrating the importance of the spatial-temporal correlations in point cloud video learning.

\begin{figure}[tbp]
\begin{center}
\includegraphics[width=0.9\linewidth]{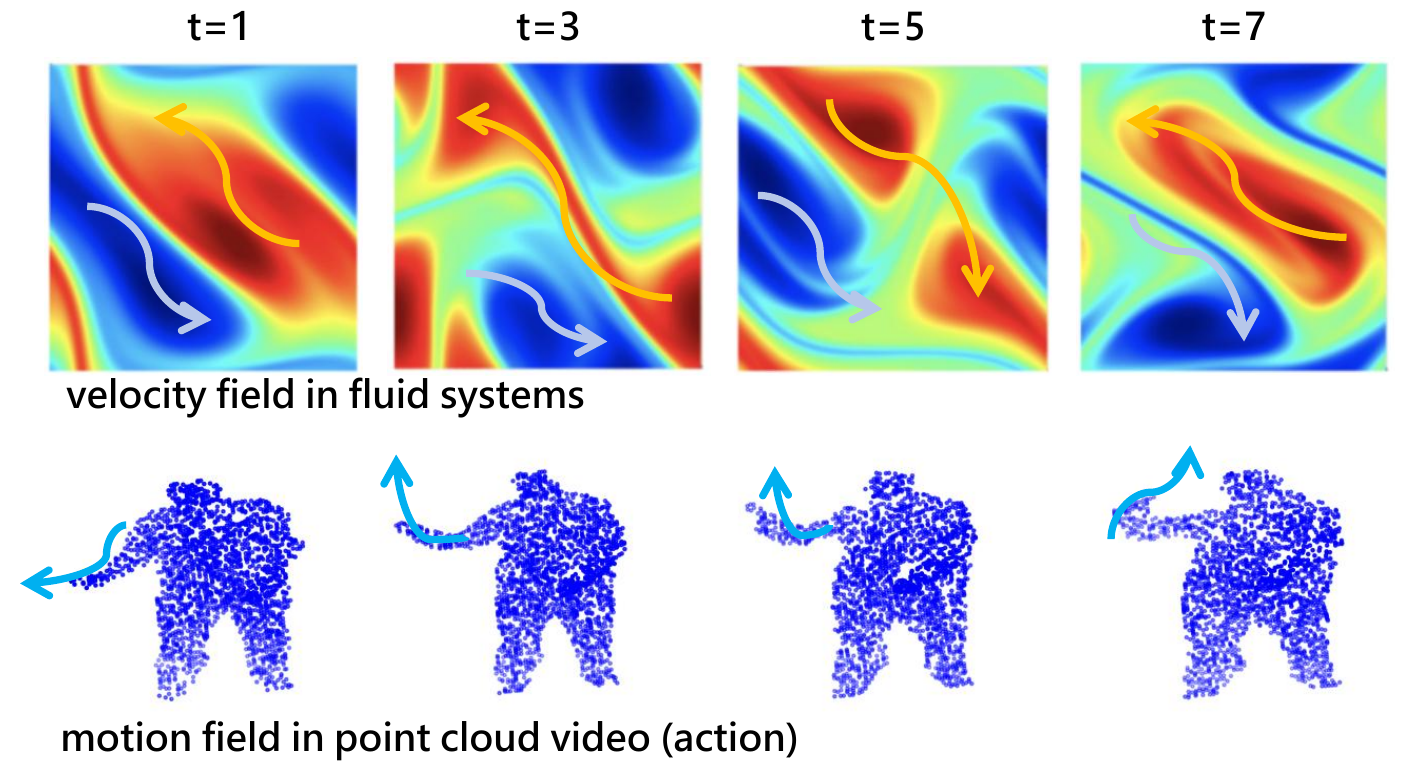}
\end{center}
\vspace{-0.3cm}
\caption{Motivation 2: Developing a PDE system for point cloud video learning is inspired by the analogy between velocity field learning in fluid systems (top figure from \cite{brunton2023machine}) and motion field learning in point cloud videos (bottom). In practice, PDEs are usually discretized into high-dimensional coordinate spaces, as illustrated by the successive grid frames in the top figure.}
\label{fig:motivation}
\vspace{-0.3cm}
\end{figure}

For the second point, previous methods for traditional video data with grid-based structure often model the pixel-wise optical flow \cite{Sun_2018_CVPR, wang2024motion,app13148003, Ding_Wang_Zhou_Shi_Lu_Luo_2020} to provide spatial-temporal correlation in the video data. However, in point cloud video data, the uncertainty in frame-to-frame correlation results in points appearing inconsistently across different frames. 
This inconsistency, combined with the absence of explicit RGB features, makes point-wise tracking challenging and optical flow unreliable. In this context, our work on point cloud video analysis draws inspiration from previous PDE-based methods that have explored applications on irregular data, such as in solid and fluid analyses, as seen in studies like \cite{fanaskov2023spectral, wu2023LSM, liu2023htnet}. Analogically, these systems aim to study the deformation of spatial points over time due to motion, conceptualizing it as a unified representation across spatial and temporal dimensions. This representation is akin to the velocity field in fluid systems and the motion field in point cloud videos (see Figure \ref{fig:motivation}). Drawing on techniques used in fluid dynamics, we aim to develop PDE models tailored for point cloud video learning. Importantly, following recent advances such as Latent Spectral Models \cite{wu2023LSM}, PDE-style structures are adopted here as functional priors operating in the representation space, without requiring a literal physical interpretation. While human behavior in point cloud videos is not governed by physical laws as in fluid systems, we adopt PDEs as a mathematical abstraction to regularize intrinsic spatio-temporal variation, independent of specific physical drivers.

Despite the discreteness of point cloud data, such an analogy remains valid, as PDEs are commonly discretized when applied to physical systems (e.g., grids in computational fluid dynamics, see Figure \ref{fig:motivation}). However, existing PDE models cannot be seamlessly integrated with point cloud video, because the tasks in the physical domain differ fundamentally from typical tasks like classification in point cloud video. The PDE system must be adapted specifically for point cloud video understanding. Additionally, the irregular motions in point cloud videos add another layer of complexity, necessitating the exploration of suitable solutions.

To our knowledge, we are the first to explore the application of PDE models to point cloud video analysis. Prior to this, several standard methods have been developed to process point cloud video data directly. For instance, the PSTNet model family \cite{fan2021pstnet, fan2022pstnet2, fan21p4transformer, fan2023psttransformer} first applies spatial convolution to the spatial points and then sequentially constructs point tubes over time using temporal convolutions. The other two-stream methods \cite{Zhong2022kinet, liu2021GeometryMotionnet, liu2022GeometryMotionTransformer} process spatial and temporal features in separate streams. Instead of enhancing spatial-temporal correlations, these methods encounter conflicts between spatial and temporal dimensions, where spatial irregularities negatively impact temporal modeling, and vice versa. While existing methods lack explicit regularization, our approach introduces a PDE-based module that guides the learning of such spatial-temporal correlations. We seek to harmonize the interactions between spatial and temporal dimensions, thereby improving performance in point cloud video understanding.

We now introduce our MotionPDE. We formulate a PDE for modeling spatio-temporal correlations in point cloud video data, inspired by the classical Navier-Stokes equation typically used in fluid systems. Unlike fluid deformation, which involves physical factors like density, pressure, and viscosity, point cloud deformation is influenced primarily by non-physical factors. Therefore, excluding those physical factors, we simplify the equation to $\partial \mathbf{u} / \partial t - (\mathbf{u} \cdot \nabla) \mathbf{u}=0$, where the first term denotes the \textbf{temporal} variations in point cloud video, while the second term captures the influence of \textbf{spatial} representation on itself. The solving target of this PDE is to harmonize the spatial and temporal terms. To achieve this, we design a contrastive learning structure to guide the PDE-solving process, ensuring a balanced integration of spatial and temporal information. 

Specifically, given that the backbone features from a point cloud video model such as PSTNet \cite{fan2021pstnet}, which already encapsulate joint spatial-temporal information, we apply separate pooling operations along the spatial and temporal dimensions to generate two complementary feature representations. These representations are then used to initialize the spatial and temporal terms in our simplified PDE formulation. We then combine the attention mechanism and the operator-learning model equipped with the classic trigonometric spectral method \cite{tolstov2012fourier} to model the spatial $(\mathbf{u} \cdot \nabla) \mathbf{u}$ and temporal $\partial \mathbf{u} / \partial t$ terms. Finally, a contrastive loss is adopted to optimize and harmonize the spatial and temporal terms. 

\begin{figure}[tbp]
\begin{center}
\includegraphics[width=\linewidth]{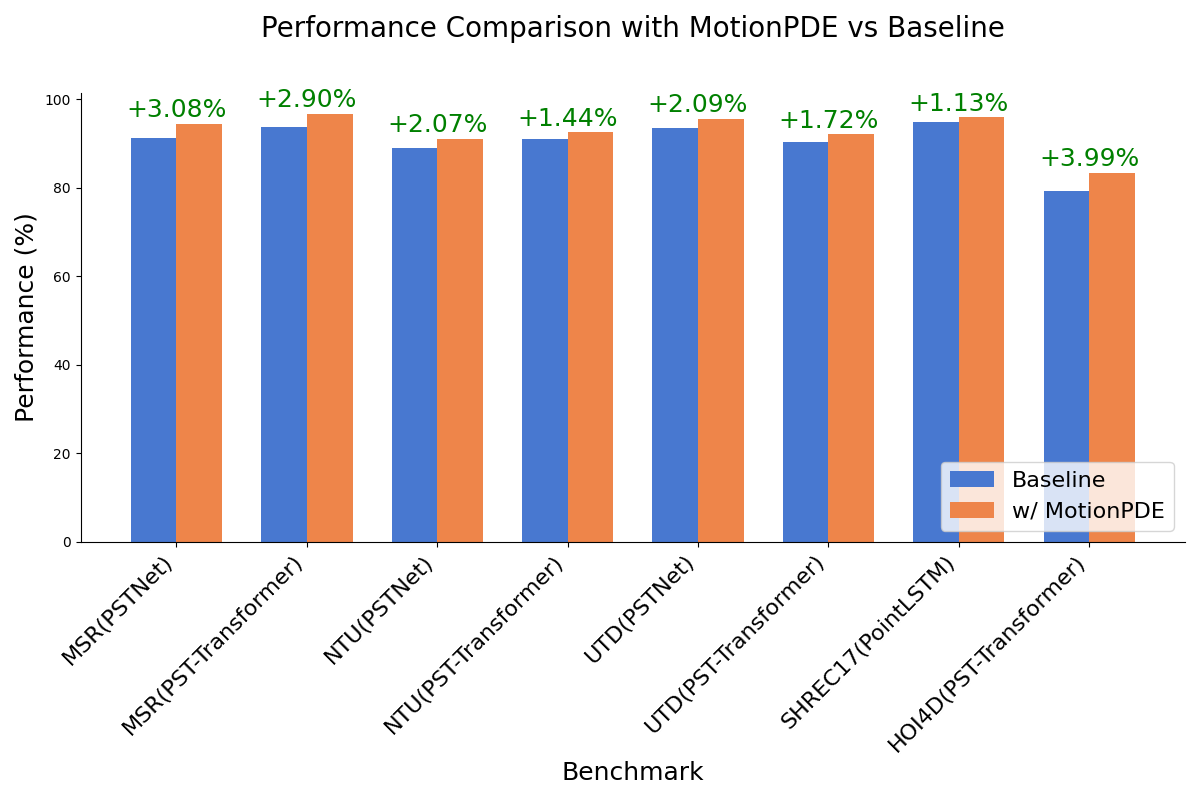}
\end{center}
\vspace{-0.3cm}
\caption{Performance boots with our MotionPDE.}
\label{fig:performance_intro}
\vspace{-0.3cm}
\end{figure}

MotionPDE unleashes the great potential of the existing backbones, as shown in Figure \ref{fig:performance_intro}. As a plug-and-play module, our design sustains a lightweight framework with minimal parameters and computational complexity. Our MotionPDE functions as a regularization by supervising the backbone through backward propagation, which forces the backbone to learn spatial-temporal correlations in point cloud video under the PDE guidance better. Furthermore, we explore the self-supervised potential of MotionPDE by providing pre-trained models. Our results demonstrate that the spatial-temporal correlations learned by MotionPDE are easily transferable across different datasets, proving that MotionPDE can be universal and dataset-independent. The experimental results substantiate that MotionPDE enhances the efficacy of extant backbone models and is instrumental in achieving state-of-the-art performance across various benchmarks in diverse downstream tasks. In summary, the contributions of this work are summarized as follows:
\begin{itemize}
\item We propose a novel perspective that views the spatial-temporal correlations learning of point cloud video as a PDE-solving problem. The innovation lies in constructing a PDE system inspired by the classical Navier-Stokes equations, leveraging the parallels between point cloud video and fluid systems. Our MotionPDE acts as a regularizer for the backbone, compelling it to learn improved spatio-temporal correlations in point cloud videos under the guidance of PDE.
\item Our MotionPDE is tailored for point cloud video understanding and can be seamlessly integrated as a plug-and-play enhancement module for existing backbones. Additionally, it can function as a self-supervised structure, offering pre-trained weights that are applicable across various datasets.
\item With extensive experiments on multiple benchmarks, we demonstrate that MotionPDE elevates baseline performance to a state-of-the-art level. We show that the spatial-temporal correlations of point cloud video learned by MotionPDE are dataset-independent and task-independent, making them easily transferable across different scenarios. 
\end{itemize}

\section{Related Works}

\subsection{PDE Models for Sequential Data}

\subsubsection{Application of PDEs} Partial differential equations (PDEs) are prevalent in scientific computing, mathematics, and engineering. They typically involve multiple independent variables, which may include spatial variables alone or both spatial variables and time. Therefore, they have long been used in sequential data for spatial-temporal correlation learning. PDE models have been adopted in the scientific domain, including weather modeling, manufacturing, solid and fluid analyses, \textit{etc.} \cite{pdeStan, verma2024climode, fanaskov2023spectral, wu2023LSM, liu2023htnet}. Additionally, PDE has garnered interest in vision research, finding applications in tasks like image processing \cite{pdevision, mang2018pde, zhang2023pdeevent, Boquet2023pdeopticalflow}, point cloud compression \cite{yang2023pde}, and video prediction \cite{guen2020disentangling, wu2023predictive}. 

\subsubsection{Solving Methods for PDEs} While the PDE-solving problems have been widely explored with spectral methods \cite{gottlieb1977numerical, fornberg1998practical} and numerical methods \cite{solin2005partial, grossmann2007numerical} since the last century. Recent research has explored deep learning models for PDEs due to their strong nonlinear modeling capabilities. These methods can generally be categorized into equation-constraint and operator-learning approaches. The former approach \cite{yu2018deep, zhai2023deep} parameterizes the solution to a PDE as a deep learning model and requires constraints of the equation. In the point cloud video domain, although we successfully built the PDE equation for the spatial-temporal correlation modeling, the constraints of the equation are challenging to formulate. Conversely, the latter approach \cite{li2020fourier, tran2023factorized, fanaskov2023spectral, liu2023htnet} seeks to use deep models to approximate the mapping between input-output pairs of PDE solutions without explicitly enforcing PDE constraints.

In this study, we aim to integrate PDE models into point cloud video understanding tasks. To achieve this, we have developed a novel MotionPDE model that utilizes PDEs to capture spatial-temporal correlations in point cloud video. Our approach is grounded in the operator-learning paradigm, utilizing spectral methods as the fundamental operators for solving PDEs. Additionally, we have designed a contrastive-based architecture to guide and refine the PDE-solving process without the need for explicitly enforcing PDE constraints.

\subsection{Point Cloud Video Understanding}

Point cloud video contains complex spatial-temporal information and combines an intricate structure with both unordered (intra-frame) and ordered (inter-frame) nature. Early methods either simplify their structure by dimensionality reduction using projections \cite{luo2018fast}, or adopt voxelization to construct a regulated grid-based data \cite{Minkowski, wang20203dv}. Similar to projections/voxel-based methods in static point clouds, these methods also faced information loss and issues with processing efficiency. Recent methods \cite{liu2019meteornet, Min2020pointlstm, fan2021pstnet, fan21p4transformer} tend to process the point cloud video directly using a set abstraction \cite{qi2017pointnet++} approach. For instance, \cite{fan2021pstnet} proposed a 4D convolution that implicitly captures the dynamics of adjacent point cloud frames by performing a set abstraction. The method processes the point cloud sequence recursively. An improved version published later \cite{fan2022pstnet2} proposed to enhance the capture of dynamics with an additional temporal convolution. These point-based methods focus more on the motion representation and try to improve the dynamic capture process in different ways. P4Transformer \cite{fan21p4transformer} and PST-Transformer \cite{fan2023psttransformer} capture dynamics by searching related points in the spatial-temporal space with attention-based networks. Kinet \cite{Zhong2022kinet} proposes a kinematics-inspired neural network and models the dynamics in point cloud sequences using scene flow. Our MotionPDE is built upon a brand-new perspective that regularizes the motion representation learning in point cloud video with a PDE model. The proposed method is designed to enhance the existing backbone models in a plug-and-play fashion, while also facilitating self-supervised learning.

\begin{figure*}[tbp]
\begin{center}
\includegraphics[width=\linewidth]{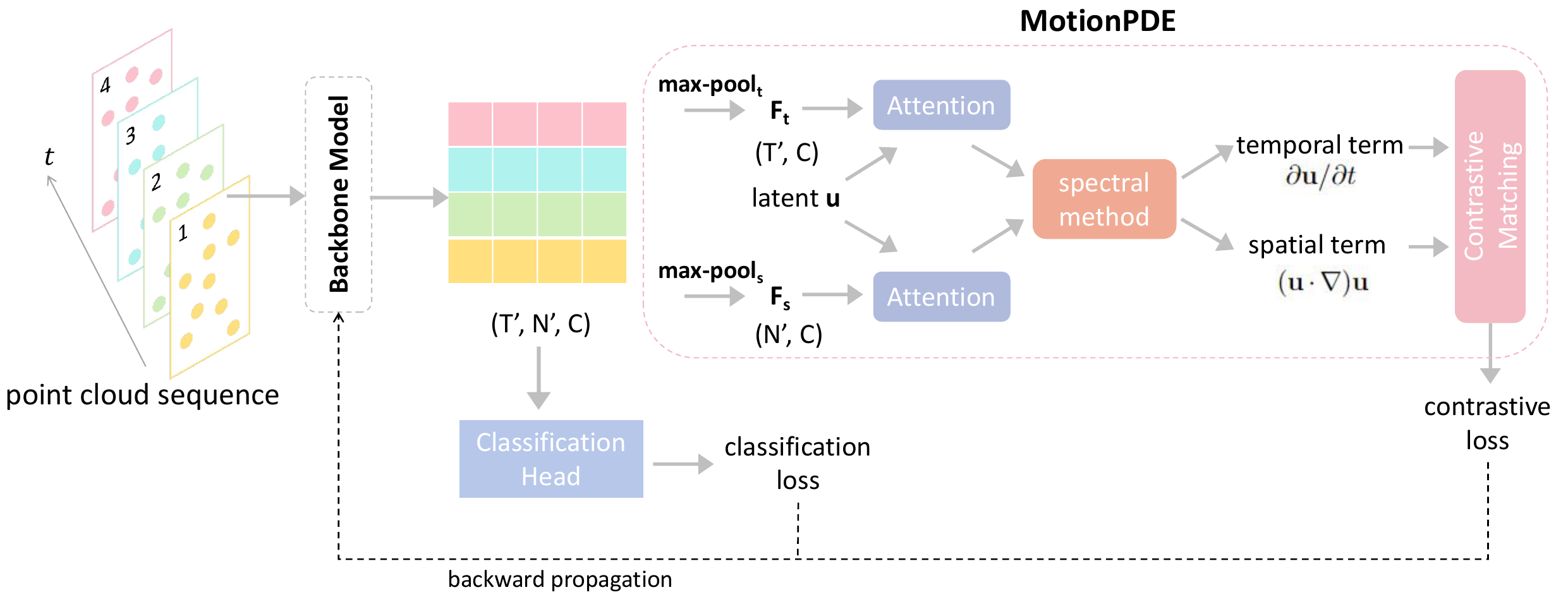}
\end{center}
\vspace{-0.3cm}
\caption{Overall architecture. MotionPDE applies separate pooling operations along the spatial and temporal dimensions to construct two complementary feature streams, enabling independent modeling of spatial structure and temporal dynamics. Then, an attention layer is adopted to project features into the latent space. A latent $\mathbf{u}$ is randomly initialized and serves as the query, while the decomposed features, $F_s$ and $F_t$, both serve as keys and values. Following a spectral method core, the output is then fed into a contrastive matching loss. The output loss provides additional supervision, compelling the backbone to learn better spatial-temporal correlations in point cloud videos guided by our MotionPDE.}
\label{fig:motionpde}
\vspace{-0.3cm}
\end{figure*}

\section{Proposed Method}

Figure \ref{fig:motionpde} illustrates the mechanism of the proposed MotionPDE. MotionPDE can be integrated with current mainstream backbones for point cloud video analysis. Our method is designed to regularize spatial-temporal correlation learning of the backbone models to boost the baseline performance. The spatio-temporal knowledge learned through our MotionPDE will optimize the learning weights of the backbone through backward propagation. We design two training strategies for our methods. The first strategy optimizes the backbone together with MotionPDE in an end-to-end manner. The second strategy employs a self-supervised approach to provide a pre-trained model using MotionPDE. 

In the following sections, we first explain the construction and the core solving process for the spatial-temporal PDE in point cloud video. Then, we illustrate the proposed MotionPDE module in detail. Finally, we detail the training methods for our model. We start with some basic concepts and symbol definitions, which will help with comprehension in the following reading:
\begin{itemize}
\item $\mathbf{u}$ and $\mathbf{u}_0$: both represent the motion field in point cloud video, which could be fine-tuned and updated by the PDE model, and $\mathbf{u}_0$ represents its initial state.
\item $\mathbf{u}(\mathit{l})$: equivalent to $\mathbf{u}$, represents the solution of the PDE system.
\item $\mathcal{M}(\mathbf{u})$: $\mathcal{M}$ represents the PDE model used to update the motion field $\mathbf{u}$, and $\mathcal{M}(\mathbf{u})$ is defined as the process of approximating the solution of the PDE system. 
\item A PDE system: a general concept that represents our MotionPDE method. Its components include the PDE model, the motion field in point cloud video, and the solution target, guided by the PDE equation.
\end{itemize}

\subsection{Constructing Spatial-Temporal PDE in Point Cloud Video}

Given a point cloud with $N$ points presented as $P = \{p_1, p_2, ..., p_N\}$, where each point $p_i \in \mathbb{R}^3$ is specified by the geometric coordinates $\{x, y, z\}$, a point cloud video containing $T$ frames of point clouds can be presented as $V = \{P_1, P_2, ..., P_T\}$. This combines the unordered nature of intra-frame data with the ordered sequence of inter-frame data. A temporal sequence $V$ can be viewed as sampling from an underlying continuous spatio-temporal field. Under this perspective, each point $p_i(t) \in P_t$ follows an implicit trajectory governed by a latent velocity field,
\begin{equation}
\frac{d p_i(t)}{dt} = \mathbf{u}(p_i(t), t),
\end{equation}
which abstracts how geometry evolves over time. To model the motion information conveyed by the spatial points in the point cloud video data, we construct our guiding equation, which is derived from the classical Navier-Stokes equation:
\begin{equation}
    \partial \mathbf{u} / \partial t - (\mathbf{u} \cdot \nabla) \mathbf{u}=0. \label{eq:ns}
\end{equation}
It is composed of a temporal derivative $\partial \mathbf{u} / \partial t$ and a convection term $(\mathbf{u} \cdot \nabla) \mathbf{u}$ to model the nonlinear motion of the point cloud. Although the physical laws may not apply to the motion on point cloud video, we make a virtual velocity field $\mathbf{u}$ to represent the motion field \cite{song2022pref} in a point cloud video. Here, the PDE formulation serves as a mathematical device for modeling the temporal evolution of geometry, without implying physical equivalence. This equation describes the dynamic motion of the point cloud video with a motion field $\mathbf{u}$ that captures the motion of an object or scene in a point cloud. Note that Equation \ref{eq:ns} is similar to the classical Euler equation,\footnote{The Euler equation is a special case of the Navier–Stokes equation for inviscid fluids, written as \(\partial \mathbf{v} / \partial t + (\mathbf{v} \cdot \nabla) \mathbf{v} = 0\), where \(\mathbf{v}\) represents the velocity field. The Euler equation and Equation \ref{eq:ns} are mathematically interchangeable through the variable substitution $\mathbf{u} = -\mathbf{v}$.} but we intentionally adopt a sign-inverted formulation. This transformation is mathematically equivalent via a variable substitution and allows the spatial and temporal terms to be aligned directly \(\partial \mathbf{u} / \partial t = (\mathbf{u} \cdot \nabla) \mathbf{u}\) rather than inversely \(\partial \mathbf{u} / \partial t = -(\mathbf{u} \cdot \nabla) \mathbf{u}\). Such alignment better suits our contrastive learning framework, where paired representations are expected to be similar in direction and meaning.

To solve this PDE, we let $\mathit{l} \in \Omega \subset \mathbb{R}^d$ represent the spatio-temporal location and the $\mathbf{u}(\mathit{l}) \in \mathbb{R}, \forall \mathit{l} \in \Omega$ represent the unknown solution of the PDE system. The solution of the PDE can be formulated as follows:
\begin{gather}
    \mathcal{F}(\mathit{l}, t) \xrightarrow{\mathcal{M}} \mathbf{u}(\mathit{l}), \label{eq:backbone}\\
    \nonumber \mathit{l} = P_t, ~ t=1,2,3,...,T.
\end{gather}
$P_t$ represents a point cloud frame at time $t$ and $\mathcal{F}$ is a chosen backbone models, e.g., PSTNet \cite{fan2021pstnet} and PSTTransformer \cite{fan2023psttransformer} that generates feature embedding of the point cloud video. $\mathbf{u}(\mathit{l})$ satisfies $ \partial \mathbf{u} / \partial t - (\mathbf{u} \cdot \nabla) \mathbf{u}=0$.

While the models: $\mathcal{F}$ and $\mathcal{M}$ are optimized simultaneously, the solution target of our PDE system is to optimize the learning parameters in the model $\mathcal{F}$ so as to satisfy Equation \ref{eq:ns}. In other words, our PDE model functions as a regularizer that forces the backbone $\mathcal{F}$ to learn the spatial-temporal correlations in point cloud video under the PDE guidance better.

\subsection{The MotionPDE Module}
\label{sec:motionpdemodule}

Given the solution process in Equation \ref{eq:backbone}, guided by PDE in Equation \ref{eq:ns}, we now need to identify a supervision target for the model learning. With a simple transpose, the Equation \ref{eq:ns} turns to:
\begin{equation}
    \partial \mathbf{u} / \partial t = (\mathbf{u} \cdot \nabla) \mathbf{u}. \label{eq:nstranspose}
\end{equation}

This motivates us to build a contrastive learning structure and to model the temporal derivative and the convection terms \textbf{separately}. 

\subsubsection{Vanilla PDE Model}
\label{sec:vanillapde}

Due to the inherent complexity of our problem and the lack of suitable references from prior studies, it is challenging to define appropriate boundary conditions. Consequently, traditional numerical methods \cite{gottlieb1977numerical, solin2005partial} are not directly applicable to solving this PDE. To address this, we draw upon the operator-learning paradigm using deep models, which is theoretically supported by classical approximation theory. Specifically, the universal approximation theorem for nonlinear operators \cite{chen1995universal} provides a foundation for using neural networks to approximate the input-output mappings defined by PDEs. Building on this, we follow recent advances \cite{liu2023htnet,li2020fourier, fanaskov2023spectral, karniadakis2021physics} that use deep models to directly learn such mappings, without requiring explicitly defined boundary conditions. The vanilla PDE model is composed of a pooling operation, the multi-head cross-attention, and further enhanced with the classic spectral method \cite{tolstov2012fourier} in the next section.

Both the derivative along the temporal dimension $\partial t$ and the derivative along the spatial dimension $\nabla \mathbf{u}$ aim to capture the rate of change. With similar dimensionality reduction operations, we utilize different max-pooling with an attention layer on the backbone feature to capture significant changes along the temporal and spatial dimensions separately. 

Given the backbone feature $\in \mathbb{R}^{T' \times N' \times C}$, where $N'$ is the number of aggregated features after the spatial stride, $T'$ is the number of aggregated features after the temporal stride, and $C$ is the number of feature channels, we utilize max-pooling along the spatial and temporal dimensions of the backbone feature separately to generate $F_s \in \mathbb{R}^{N' \times C}$ and $F_t \in \mathbb{R}^{T' \times C}$ (see Figure \ref{fig:motionpde}). Then, an attention layer is used to project $F_s$ and $F_t$ into a latent space with a token number $N_\text{Token}$. The attention layer we use is a standard multi-head cross-attention (MHCA):
\begin{gather}
    \mathbf{u}_s = \text{MHCA}(\mathbf{u}, F_s), \\
    \mathbf{u}_t = \text{MHCA}(\mathbf{u}, F_t).
\end{gather}

Here the $\mathbf{u} \in \mathbb{R}^{N_\text{Token} \times C}$. The calculations for $\mathbf{u}_s$ and $\mathbf{u}_t$ are the same except for different inputs $F_s$ and $F_t$. We use $\mathbf{u}_F \in \mathbb{R}^{N_\text{Token} \times C}, ~ \text{where} ~ \mathbf{u}_F = \mathbf{u}_s ~ \text{or} ~ \mathbf{u}_t$, to represent the projected latent feature. The $\text{MHCA}(\mathbf{u}, F)$ can be formulated as follows:
\begin{equation}
    \mathbf{u}_F = \mathrm{Concat}(H_1,\dots,H_m), \label{eq:mhca2}
\end{equation}
where its constituents $H_i$ is equivalent to the attention head $\text{MHCA}(\mathbf{u}, F)_i$
\begin{equation}
    H_i = \mathrm{Softmax}(\{\mathbf{u} \cdot W_i^{Q}\} \times \frac{\{FW_i^{K}\}^{T}}{\sqrt{d}}) \times FW_i^{V}, i=1,...,m. \label{eq:mhca1}
\end{equation}
Different $W_i$ indicate the weights of the queue, key, and value, respectively. After concatenating the attention heads, we obtain the projected latent features $\mathbf{u}_s$ and $\mathbf{u}_t$. 

In our implementation, the MHCA modules used for the \(\mathbf{u}_s\) and the temporal feature sequence \(\mathbf{u}_t\) share the same architecture. Importantly, we do not apply any positional encoding to either \(F_s\) or \(F_t\). For \(F_s\), this avoids imposing artificial order on unordered point sets; for \(F_t\), temporal structure is implicitly captured by the backbone, and explicit encoding showed no consistent benefits in our preliminary experiments.

\subsubsection{Enhancing PDE Model with Spectral Method}
\label{sec:operatorlearnin}

The vanilla PDE model provides a promising solution to solving the PDE problem. However, such a design attempts to approximate the input-output mapping with a single operator, which raises the problem of maintaining network efficiency, due to the complexities of input-output mappings in high-dimensional space \cite{wu2023LSM, karniadakis2021physics}. We tackle this problem by combining the attention mechanism and the classic spectral method \cite{tolstov2012fourier} for PDE. While the attention mechanism provides a universal approximation capability, the spectral method decomposes the complex nonlinear mappings into multiple basis operators. In this way, without the necessity to build a deep attention model, we are able to use a one-layer MHCA to achieve efficient performance.

Given the projected latent feature $\mathbf{u}_F$ from the vanilla PDE model, we further update the feature with the following formulation:
\begin{equation}
    \mathbf{u}_F' = \mathcal{M}(\mathbf{u}_F),
\end{equation}
where
\begin{gather}
    \mathcal{M}(\mathbf{u}_F) = \sum_{i=1}^{O} \mathit{w}_i \mathcal{M}_{i}. \label{eq:operators1}
\end{gather}
Here $O$ is the number of operators and $\mathit{w}_i$ is a learnable weight.

For each $u_i \in \mathbf{u}_F, i = 1 ... N_\text{Token}$, we use trigonometric functions as the basis operators following \cite{lu2021learning} and \cite{li2021fourier}:
\begin{gather}
    \mathcal{M}_{(2k-1)}(u_i) = sin(ku_i) \label{eq:bias1}, \\
    \mathcal{M}_{(2k)}(u_i) = cos(ku_i) \label{eq:bias2}, 
\end{gather}
where $k \in \{1, ..., O/2\}$ and $O$ is even. Then, the calculation of the mapping output can be formulated as follows:
\begin{gather}
\label{eq:output}
\resizebox{\linewidth}{!}{$
    \mathcal{M}(\mathbf{u}_F) = \mathbf{u}_F + \mathit{w}_{sin}[\mathcal{M}_{(2k-1)}(\mathbf{u}_F)]_{k=1}^{O/2} + \mathit{w}_{cos}[\mathcal{M}_{(2k)}(\mathbf{u}_F)]_{k=1}^{O/2}.
$}
\end{gather} 
Both $\mathit{w}_{sin}$ and $\mathit{w}_{cos}$ are learnable papemeters.

By a process of combining the pooling methods and the multi-head cross-attention elaborated in Section \ref{sec:vanillapde}, our PDE model $\mathcal{M}$ is able to approximate the PDE solution effectively. We officially formulated the process as follows:
\begin{equation}
    \mathcal{M}(\mathbf{u}, F), F = F_s 
 \ \text{or} \ F_t.
\end{equation}

\subsubsection{The Contrastive Loss}

We can now construct the spatial-temporal PDE in Equation \ref{eq:ns} as:
\begin{gather}
    \partial \mathbf{u} / \partial t - (\mathbf{u} \cdot \nabla) \mathbf{u}=0, \\
    \mathcal{M}_T(\mathbf{u}, F_t) - \mathcal{M}_S(\mathbf{u}, F_s) = 0.
\end{gather}
The suffixes $T$ and $S$ are used to distinguish between different terms.

Then a variant of the contrastive InfoNCE loss \cite{oord2018representation} is used to supervise the temporal derivative $\mathcal{M}_T$ and the convection terms $\mathcal{M}_S$. The loss function can be formulated as follows:

\begin{equation}
\label{eq:contrastive_s}
    \mathbf{S} = \left( \mathcal{M}_T \cdot \mathcal{M}_S^\top \right) / \tau,
\end{equation}
\begin{equation}
    \mathcal{L}(\mathbf{S}) = \sum_{i=1}^{N_\text{Token}} -\log \left( \frac{\exp(\mathbf{S}_{i, i})}{\sum_{j=1}^{N} \exp(\mathbf{S}_{i, j})} \right),
\end{equation}
\begin{equation}
    \mathcal{L}_{\text{InfoNCE}} = \lambda\frac{1}{2} \left( \mathcal{L}(\mathbf{S}) + \mathcal{L}(\mathbf{S}^\top) \right),
\end{equation}
where $\tau$ is a temperature that controls the network sensitivity to positive and negative samples, and $\lambda$ is the loss weight. 

In Equation \ref{eq:contrastive_s}, we perform a reshape operation that flattens the token across the batch dimension to enable cross-sample token-level negative sampling. In this contrastive loss, we treat each token pair $(\mathcal{M}_s^i, \mathcal{M}_t^i)$ as a positive pair, while all non-corresponding token pairs $(i \ne j)$ are treated as negatives. This is a well-established strategy in contrastive learning frameworks like SimCLR \cite{chen2020simple}, where negatives do not rely on explicit physical dissimilarity. This design encourages one-to-one alignment between spatial and temporal views, thereby optimizing and harmonizing their representations in the shared latent space.

Overall, we show the hyperparameters of our MotionPDE as follows:
\begin{itemize}
\item Hidden dimension in the attention layer: 256 or 1024 (for different implementations)
\item Head number for the multi-head cross-attention: 8
\item Latent token number $N_\text{Token}$: 4
\item Number of operators $O$: 16
\item Temperature $\tau$: 0.1
\item Loss weight $\lambda$: 0.1
\end{itemize}
Those settings are carefully fine-tuned on the MSRAction-3D implementation and followed in other experiments. The best settings of our MotionPDE may vary for different benchmarks, but we decided not to perform excessive fine-tuning and show the generic enhancement achievable by our MotionPDE.

\subsection{Training Strategy}
Our MotionPDE works as a plug-and-play enhancement module and can be trained end-to-end (as shown in Figure \ref{fig:motionpde}), or in a pre-train \& fine-tune manner utilizing a self-supervised approach.

\subsubsection{End-to-End Training} As shown in Figure \ref{fig:motionpde}, our MotionPDE works as a plug-and-play enhancement module to the backbone. We use both the contrastive loss and the classification loss to supervise the network learning. In this way, the backbone tends to satisfy the downstream tasks with additional regularization under the PDE guidance.

\begin{table*}[htbp]
\caption{Action recognition results compared to state-of-the-art methods. * represent the results obtained using our own implementation.}
\vspace{-0.3cm}
\label{tab:actrecognition}
\begin{center}
{
\begin{tabular}{lccccccc}
\toprule
    \multirow{2}{*}{Methods} & \multirow{2}{*}{Reference} & \multirow{2}{*}{MSR}  & \multicolumn{2}{c}{NTU60} & \multicolumn{2}{c}{NTU120} & \multirow{2}{*}{UTD} \\
    & & & Cross-Subject & Cross-View & Cross-Subject & Cross-Setup\\
\midrule
    MeteorNet \cite{liu2019meteornet} & ICCV 19 & 88.50 & - & -& - & - & - \\
    3DV-PointNet++ \cite{wang20203dv} & CVPR 20 & - & 88.8 & 96.3 & 82.4 & 93.5 & - \\
    P4Transformer \cite{fan21p4transformer} & CVPR 21 & 90.94 & 90.2 & 96.4 & 86.4 & 93.5 & - \\
    \rowcolor{cyan!20} PSTNet \cite{fan2021pstnet} & ICLR 21 & 91.20 & 90.5 (88.83*) & 96.5 & 87.0 (86.33)	& 93.8 & 93.49* \\
    PointMapNet \cite{pointmapnet} & Symmetry 23 & 91.91 & 89.4 & 96.7 & - & - & 91.61 \\
    SequentialPointNet \cite{li2021sequentialpointnet} & IEEE TII 22 & 91.94 & 90.3 & 97.6 & 83.5 & 95.4 & 92.31 \\
    3DInAction \cite{ben20233dinaction} & CVPR 23 & 92.23 & 89.3 & - & - & -  & - \\
    HyperPointNet \cite{Li2022Hyper} & ICME 22 & - & 90.2 & 97.3 & 83.2 & 95.1 & - \\
    PPTr \cite{wen2022pptr} & ECCV 22 & 92.33 & - & - & - & -  & 89.07* \\
    PointCPSC \cite{sheng2023point} & ICCV 23 & 92.68 & - & - & - & -  & - \\
    PSTNet++ \cite{fan2022pstnet2} & IEEE TPAMI 22 & 92.68 & 91.4 & 96.7 & 88.6 & 93.8 & - \\
    Kinet \cite{Zhong2022kinet} & CVPR 22 & 93.27 & 92.3 & 96.4 & - & -  & - \\
    Mamba4D \cite{liu2024mamba4d} & CVPR 25 & 93.38 & - & - & - & -  & - \\
    \rowcolor{teal!15} PST-Transformer \cite{fan2023psttransformer} & IEEE TPAMI 23 & 93.73 & 91.0 & 96.4 & 87.5 & 94.0 & 90.27* \\
    KAN-HyperpointNet \cite{chen2025kanhyper} & ICASSP 25 & 95.59 &  91.6 & 98.4 & - & - & - \\
\midrule
    PPTr 
    {\textbf{MotionPDE}} & - & 93.11 & - & - & - & -  & 91.40 \\
    \rowcolor{cyan!20} PSTNet
    {\textbf{MotionPDE}} & - & 94.28 & 90.90 & 96.77 & 88.72 & 93.90 & 95.58 \\
    PSTNet++
    {\textbf{MotionPDE}} & - & 95.22 & 92.42 & 97.01 & 89.11 & 94.21 & - \\
    \rowcolor{teal!15} PST-Transformer
    {\textbf{MotionPDE}} & - & 96.63 & 92.44 & 97.24 & 89.11 & 94.42 & 91.99 \\
\bottomrule
\end{tabular}
}
\end{center}
\vspace{-0.5cm}
\end{table*}

\subsubsection{Pre-train \& Fine-tune Manner} During the pre-training, we intentionally omit the classification head to avoid using labels at this stage. The network is supervised by the contrastive loss only. Then, we only fine-tune the encoder with a classification head, excluding the MotionPDE. In this way, the model will first learn the general spatial-temporal correlations in point cloud video, and then further learn the downstream tasks in the fine-tuning stage.

\section{Experiments}

To validate the performance of our MotionPDE, we conduct extensive experiments on different benchmarks and tasks including MSRAction-3D \cite{li2010msraction}, NTU RGB+D (60 \cite{shahroudy2016ntu} and 120 \cite{liu2019ntu}), UTD-MHAD \cite{utddataset}, HOI4D \cite{hoi4d_Liu_2022_CVPR}, and SHREC 2017 \cite{de2017shrec} datasets, which cover multiple foundational tasks including Action Recognition, Action Segmentation, and Gesture Recognition.

We first report the performance of our MotionPDE on different downstream tasks with end-to-end training in Section \ref{sec:actionrecognition}, Section \ref{sec:gesture}, and Section \ref{sec:actionsegmentation}. Then, we explore the self-supervised performance with our MotionPDE using the pre-train \& fine-tune strategy in Section \ref{sec:pretrain}, in which we also explore the cross-dataset pre-training that helps improve the performance of downstream tasks. We then discuss the model efficiency and the choice of hyperparameters in Section \ref{sec:param}, followed by the ablation study in Section \ref{sec:ablation}. Further analysis and visualization results are in Section \ref{sec:analysis}. The implementation setting may vary from case to case. For the hyper-settings relating to our MotionPDE module, we only fine-tune carefully in the MSRAction-3D experiments and maintain the same settings on other datasets and tasks. We did all our experiments on 80Gb NVIDIA A100 GPUs. We use 4 $\times$ A100 GPUs for the NTU RGB+D datasets \cite{shahroudy2016ntu, liu2019ntu} and a single A100 GPU for other experiments. The setting details are in Section \ref{sec:motionpdemodule}.

Since we focus on methods that directly process point cloud videos, we have chosen two backbones that effectively represent the convolution-based and transformer-based models—PSTNet \cite{fan2021pstnet} and PST-Transformer \cite{fan2023psttransformer}—as the main baselines to test our MotionPDE. We also implement MotionPDE on additional backbones like PointLSTM \cite{Min2020pointlstm}, PSTNet++, \cite{fan2022pstnet2}, and PPTr \cite{wen2022pptr}. To comprehensively demonstrate the generalizability of MotionPDE across different baselines, we include results from our own reimplementations when the original papers do not provide the corresponding numbers. These results are produced under consistent experimental settings, allowing for fair and direct comparisons with and without MotionPDE.

\subsection{Action Recognition}
\label{sec:actionrecognition}

For 3D action recognition, we perform our method on the MSRAction-3D \cite{li2010msraction}, NTU RGB+D (60 \cite{shahroudy2016ntu} and 120 \cite{liu2019ntu}), and UTD-MHAD \cite{utddataset} datset. Following the original setting in the baseline methods \cite{fan2021pstnet, fan2023psttransformer}, we split the input video into several overlapping clips and randomly sampled 2048 points for each input frame. The clip length is set to 24, with a frame sampling
stride of 1 on MSRAction-3D and 2 on NTU-RGBD and UTD-MHAD. We use the default split \cite{li2010msraction, shahroudy2016ntu, liu2019ntu, utddataset} for training and evaluation. The results compared to prior works are shown in Table \ref{tab:actrecognition}. We report the mean of clip-level prediction as the final result, as the baseline did.

\subsubsection{Implementation details for experiments with \textbf{MSRAction-3D}} 
\label{sec:msraction3d}

The dataset for action recognition includes 567 depth map sequences of 20 action classes performed by 10 subjects. We generate point cloud videos from the original depth data with the standard script in \cite{liu2019meteornet}. We train the model for 45 epochs with a batch size of 32, optimizing it with an initial learning rate of 0.05 (for PSTNet) and 0.03 (for PST-Transformer). We use the SGD optimizer with a momentum of 0.9 and a weight decay of 0.0001. Additionally, we apply a linear learning rate scheduler with 10\% warm-up steps. 

Our report demonstrates that the proposed MotionPDE significantly enhances the baseline models. Specifically, it boosts the PSTNet performance to 94.28\% accuracy and the PST-Transformer performance to 96.63\% accuracy, both achieving approximately a 3.0\% increase in accuracy.

\subsubsection{The implementation details for experiments on \textbf{NTU RGB+D}} 

\begin{table}[htbp]
\caption{The action recognition results compared to other enhancement methods on MSRAction-3D.}
\label{tab:otherenhancement}
\centering
{
\begin{tabular}{lc}
\toprule
Methods & Accuracy(\%) \\
\midrule
PPTr \cite{wen2022pptr} & \transparent{0.5}{92.33} \\
{\quad +LeaF \cite{liu2023leaf}} & {\quad + 1.51} \\
{\quad +X4D-SceneFormer \cite{jing2024x4d}} & {\quad + 1.57} \\
{\quad +C2P \cite{zhang2023c2p}} & {\quad + 2.43} \\
P4Transformer \cite{fan21p4transformer} & \transparent{0.5}{90.94} \\
    {\quad + Mamba4D \cite{liu2024mamba4d}} &  {\quad + 2.16} \\
PST-Transformer (small) & \transparent{0.5}{93.03} \\
    {\quad + Mamba4D \cite{liu2024mamba4d}} &  {\quad + 0.35} \\
\midrule
PSTNet \cite{fan2021pstnet} & \transparent{0.5}{91.20} \\
    {\quad + PointCPSC \cite{sheng2023point}} &  {\quad + 1.48} \\
    {\quad + \textbf{MotionPDE}} &  {\quad + \textbf{3.02}} \\
PST-Transformer \cite{fan2023psttransformer} & \transparent{0.5}{93.73} \\
    {\quad + \textbf{MotionPDE}} &  {\quad + \textbf{2.90}} \\
\bottomrule   
\end{tabular}
}
\vspace{-0.3cm}
\end{table}

The dataset has two versions. The NTU 60 \cite{shahroudy2016ntu} contains 60 action classes and 56,880 video samples, which is a large-scale dataset consisting of complex scenes with noisy background points. The NTU 120 \cite{liu2019ntu} is an extended version with 120 classes and an additional 57,600 video samples. For both NTU 60 and NTU 120, we train our model for 20 epochs with a batch size of 32, optimizing it with an initial learning rate of 0.05. Other settings are consistent with the implementation of MSRAction-3D.

Due to the inherent randomness, we fail to reproduce some results of the original PSTNet on the NTU RGB+D dataset. In order to fairly evaluate the performance improvement brought by our MotionPDE, we choose to report results based on our own reimplementation (the number in parentheses in Table \ref{tab:actrecognition}) of the baseline under consistent settings. To maintain transparency and acknowledge the original contributions, we also include the official results from the original PSTNet paper \cite{fan2021pstnet} for reference. Consistently, our MotionPDE brings genuine improvements to the baselines under the same settings across all reported metrics. Compared with both the PSTNet and PST-Transformer baselines, we observed an approximate 1.5\%-2\% accuracy boost in the cross-subject setting for both NTU 60 and NTU 120. The experiments in other settings also validate its effectiveness.

\subsubsection{The implementation details for experiments on \textbf{UTD-MHAD}} The dataset contains 27 classes and 861 videos with relatively long action-recognition sequences. We train the model for 20 epochs with a batch size of 32, optimizing it with an initial learning rate of 0.05. Other settings are consistent with the implementation of MSRAction-3D.

Since the baseline does not provide the results, we perform the baseline on UTD-MHAD and report the results using our own implementation. All training settings are consistent between the baseline and MotionPDE. Our findings show that PSTNet surpasses the previous SOTA method, achieving 93.49\% accuracy. Despite the limited room for improvement, our MotionPDE still brings a 2.09\% accuracy boost to the PSTNet baseline. Although the PST-Transformer performs slightly lower than previous methods, we did not fine-tune the training settings further, yet we still demonstrate the effectiveness of MotionPDE with a 1.72\% accuracy improvement.

\begin{figure}[htbp]
\begin{center}
\includegraphics[width=\linewidth]{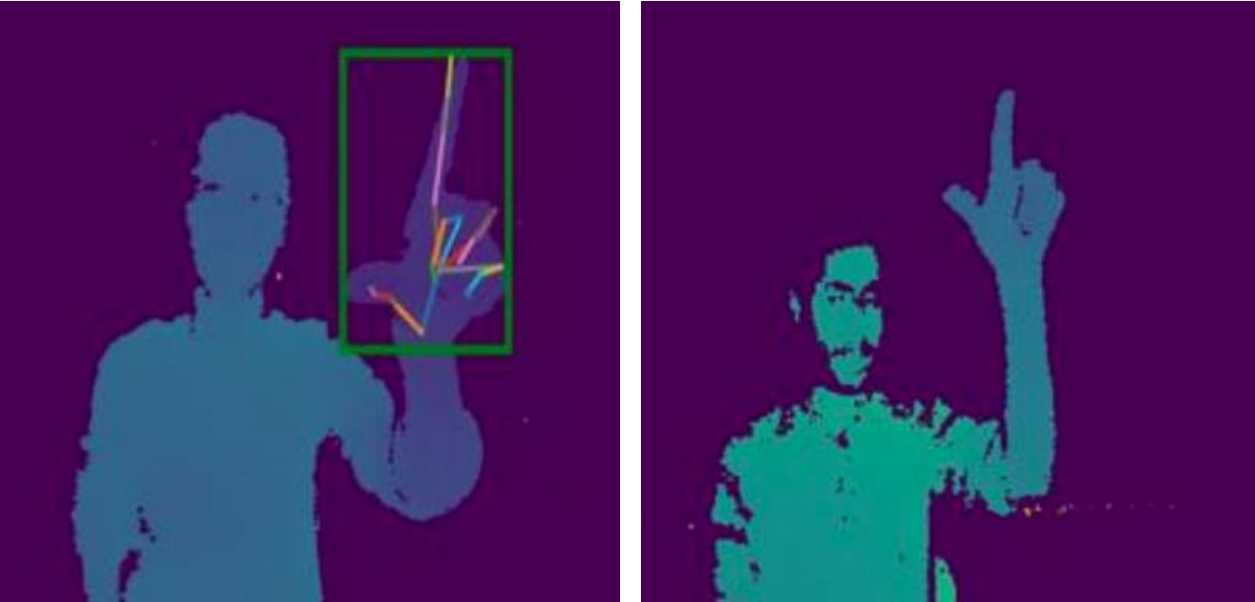}
\end{center}
\vspace{-0.3cm}
\caption{Data example of the SHREC 2017 dataset. Left: depth map with the BBox. Right: depth map without the BBox. Figure from \cite{Zhong2022kinet}.}
\label{fig:shrecbbox}
\end{figure}

\begin{table}[htbp]
\caption{The gesture recognition results compared to state-of-the-art methods. * represent results from our own implementation.}
\label{tab:gesture}
\centering
\begin{tabular}{lcc}
\toprule
Methods & BBox & Accuracy(\%) \\
\midrule
PST-Transformer (small) \cite{fan2023psttransformer}& \Checkmark & 90.47* \\
PointLSTM \cite{Min2020pointlstm} & \Checkmark & 94.7 \\
Kinet \cite{Zhong2022kinet} & \Checkmark & 95.2  \\
\midrule
PointLSTM \cite{Min2020pointlstm} & \XSolidBrush & 92.50* \\
Kinet \cite{Zhong2022kinet} & \XSolidBrush & 94.1\\
\midrule
PST-Transformer (small) \\
    {\quad + \textbf{MotionPDE}} & \Checkmark & 91.90\\
PointLSTM \\
    {\quad + \textbf{MotionPDE}} & \Checkmark & 95.83 \\
PointLSTM \\
    {\quad + \textbf{MotionPDE}} & \XSolidBrush & 93.69 \\
\bottomrule   
\end{tabular}
\vspace{-0.3cm}
\end{table}

\subsubsection{A comparison with other enhancement methods}
As a plug-and-play enhancement module, our MotionPDE is effective and easy to implement. We compare our approach to other enhancement methods in Table \ref{tab:otherenhancement}. While offering only a limited scope for comparison, we nevertheless choose enhancement methods including extra key frame information learning \cite{liu2023leaf}, extra teacher branch for knowledge distillation \cite{zhang2023c2p}, and even extra training data from RGB sequences \cite{jing2024x4d}. With baseline performances both lower (PSTNet) and higher (PST-Transformer) than the PPTr \cite{wen2022pptr}, the enhancement we bring to the baselines surpasses all those methods. 

\subsection{Gesture Recognition}
\label{sec:gesture}

For 3D gesture recognition, we perform our method on the SHREC 2017 \cite{de2017shrec} dataset. The dataset contains 2800 videos in 28 classes for gesture recognition. Following previous methods \cite{Zhong2022kinet, Min2020pointlstm}, we generate point cloud sequences from depth videos using scripts in \cite{Min2020pointlstm}. Unlike the implementation for the action recognition task, the sequence frames are sampled uniformly throughout the original video in the SHREC 2017 dataset, and no voting strategy is employed for this task. The SHREC 2017 dataset also provides bounding boxes (BBox) for the hand skeletons. Following previous settings from Kinet \cite{Zhong2022kinet}, we generated two versions of the dataset: one with points sampled within the hand region, cropped from the bounding boxes (Figure \ref{fig:shrecbbox} left), and another with sampled points that include the noisy background (Figure \ref{fig:shrecbbox} right). The sequence length is set to 32, and the point number is set to 128. The default train/val split is used following previous works \cite{Zhong2022kinet, Min2020pointlstm}.

\begin{table*}[htbp]
\caption{The Action Segmentation results compared to state-of-the-art methods. * represents the results using our own implementation. We use point clouds as the only training data in our experiments. We report methods that adopt point clouds as the only training data.}
\label{tab:actionseg}
\centering
\begin{tabular}{lccccc}
\toprule
Methods & Accuracy(\%) & Edit & F1@10 & F1@25 & F1@50\\
\midrule
MS-TCN \cite{farha2019ms} & 44.2 & 74.7 & 55.6 & 47.8 & 31.8 \\
MS-TCN++ \cite{msplusplus} & 42.2 & 75.8 & 54.7 & 46.5 & 30.3 \\
Asformer \cite{chinayi_ASformer} & 46.8 & 80.3 & 58.9 & 51.3 & 35.0 \\
P4Transformer \cite{fan21p4transformer} & 71.2 & 73.1 & 73.8 & 69.2 & 58.2 \\
\rowcolor{cyan!20} PPTr \cite{wen2022pptr} & 77.4 & 80.1 & 81.7 & 78.5 & 69.5 \\
    {\quad + Leaf} \cite{liu2023leaf} & 79.4 (+2.0) & 83.9 & 85.0 & 81.9 & 73.3 \\
    {\quad + C2P} \cite{zhang2023c2p}& 81.1 (+3.7) & 84.0 & 85.4 & 82.5 & 74.1 \\
\rowcolor{teal!15} PST-Transformer \cite{fan2023psttransformer}* & 79.22 & 59.91 & 65.39 & 62.66 & 54.64 \\
X4D-SceneFormer \cite{jing2024x4d} & 84.1 & 91.1 & 92.5 & 90.8 & 84.8 \\
    {\quad + Mamba4D \cite{liu2024mamba4d}} & 85.5 (+1.4) & 91.3 & 92.6 & 91.2 & 85.5 \\
\midrule
PPTr \\
    \rowcolor{cyan!20} {\quad + \textbf{MotionPDE}} & 82.42 (\textbf{+5.02}) & 87.24 & 88.52 & 86.62 & 80.10 \\	
PST-Transformer \\
    \rowcolor{teal!15} {\quad + \textbf{MotionPDE}} & 83.21 (\textbf{+3.99}) & 76.40 & 79.39 & 77.36 & 71.00 \\
\bottomrule   
\end{tabular}
\vspace{-0.3cm}
\end{table*}

The gesture recognition task is similar to action recognition. To prove the versatility of our MotionPDE, we further implement the PointLSTM \cite{Min2020pointlstm} as the baseline. Notice that the number of sample points is much smaller than in the datasets we used in the previous sections, so we only apply the small version of PST-Transformer with the down-sample stride of 2 in the P4DConv layer \cite{fan2023psttransformer}. We train the model for 200 epochs with a batch size of 32, optimizing it with an initial learning rate of 0.001 (for PointLSTM) and 0.01 (for PST-Transformer). We use the Adam optimizer with a weight decay of 0.0001. For PST-Transformer, the same settings are used for training the baseline and the MotionPDE. The results are reported in Table \ref{tab:gesture}. 

The BBox-cropped data are used for high accuracy in existing methods. Compared to the PointLSTM baseline, our MotionPDE brings a 1.13\% accuracy boost, which outperforms the previous SOTA. It shows that the performance of the PST-Transformer baseline is much lower than that of the compared methods. Despite the low performance of the baseline, we still show consistent enhancement thanks to our MotionPDE, bringing a 1.43\% accuracy boost.

When using the raw point cloud sequences with noisy backgrounds as input, previous SOTA methods experience significant performance drops. However, despite the absence of bounding boxes, our MotionPDE consistently improves baseline performance. This demonstrates the robustness of our approach in handling noise and its ability to effectively capture relevant features in point cloud video data, even in suboptimal conditions.

\subsection{Action Segmentation}
\label{sec:actionsegmentation}
For the 3D action segmentation task, we apply our method to the HOI4D \cite{hoi4d_Liu_2022_CVPR} dataset. The dataset contains 3863 point cloud videos in 19 action categories. Each video consists of dense frame-level annotation of action classes. Given a point cloud video, action segmentation requires the model to predict the action class for each single frame. The HOI4D dataset contains video sequences of length 150. Each frame has 2048 points. While the original dataset also provides the corresponding depth map and RGB image, our model does not use extra training data but point clouds. We compare our method with other approaches that rely solely on point cloud data.

PPTr \cite{wen2022pptr} and PST-Transformer \cite{fan2023psttransformer} are selected as the baselines and the backbones of our model. We train the model for 20 epochs with a batch size of 16, optimizing it with an initial learning rate of 0.05. Following the official benchmark, the framewise accuracy, segmental edit distance, and the segmental F1 scores at the overlapping thresholds of 10\%, 25\%, and 50\% are reported. The performance is shown in Table \ref{tab:actionseg}. We report the results from the official leaderboard of the HOI4D action segmentation challenge. Compared to the baselines, our MotionPDE achieves a 5.02\% improvement in accuracy for PPTr and a 3.99\% improvement for PST-Transformer. Our model trained with MotionPDE outperforms other enhancement methods that rely solely on point cloud data, achieving state-of-the-art results in segmental edit distance and F1 scores within the bounds of fair comparison. Meanwhile, the PST-Transformer baseline performs relatively poorly on segmental edit distance and F1 scores, primarily due to the lack of positional encoding in the original PST-Transformer method \cite{fan2023psttransformer}. Despite these limitations, our MotionPDE significantly improves these metrics, bringing them to a high level.

\subsection{Self-supervised performance with PDE}
\label{sec:pretrain}

In the previous sections, we proved that our MotionPDE significantly enhances baseline performance across a variety of benchmarks with end-to-end training. In this section, we explore the self-supervised performance with our MotionPDE using the pre-train and fine-tune strategy. We perform the pre-training on both the MSRAction-3D and the NTU RGB+D datasets to validate the performance of our MotionPDE. Both the PSTNet and PST-Transformer baselines were evaluated under this framework:

\begin{itemize}
    \item \textbf{Pre-train:} We pre-train the model for 45 epochs (on MSRAction-3D) and 10 epochs (on NTU 60). The batch size is set to 32, and the initial learning rate is set to 0.03 (on MSRAction-3D) and 0.01 (on NTU 60). Other settings are consistent with the implementation of end-to-end training. After getting the pre-trained model, we fine-tune the encoder with a classification head, excluding the MotionPDE to evaluate the pre-trained model. We test our pre-trained model on both the original pre-training datasets and various other datasets. For a fair comparison, we also report the performance of the baseline without pre-training, with the same settings.
    \item \textbf{Fine-tune:} On MSRAction-3D, we fine-tune the model for 35 epochs with a batch size of 32, optimizing it with an initial learning rate of 0.01 (for PSTNet) and 0.05 (for PST-Transformer). Other settings are consistent with the implementation of end-to-end training. On the UTD-MHAD, we fine-tune the model with the same settings that are consistent with the implementation of end-to-end training.
    
\end{itemize}

\begin{table}[htbp]
\caption{The Self-supervised performance achieved by our MotionPDE. We report the results by fine-tuning the encoder with a classification head. The same settings are used for the baseline training and the fine-tuning. * represents the results obtained using our own implementation. $^{\#}$ We further test the pre-training on a dense setting of NTU RGB+D by setting the frame sampling stride to 1.}
\vspace{-0.3cm}
\label{tab:selfsupervised}
\begin{center}
\begin{tabular}{lcc}
\toprule
    \multirow{2}{*}{\shortstack{
    ~ Fine-tune ($\rightarrow$) \\ 
    Pre-train ($\downarrow$)}}
    & \multirow{2}{*}{MSR} & \multirow{2}{*}{UTD} \\ \\
\midrule
    PSTNet \cite{fan2021pstnet} baseline * & 90.57 & 93.49 \\
    PST-Transformer \cite{fan2023psttransformer} baseline * & 94.33 & 90.27 \\
\midrule
    PSTNet on \textbf{MSR} & 94.61 & 95.34 \\
    PST-Transformer on \textbf{MSR} & 97.30 & 95.11 \\
    PSTNet on \textbf{NTU60} & 90.23$\downarrow$ & 96.04 \\
    PSTNet on \textbf{NTU60} dense$^{\#}$ & 93.60 & 96.98 \\
\bottomrule
\end{tabular}
\end{center}
\vspace{-0.3cm}
\end{table}

The results are reported in Table \ref{tab:selfsupervised}. We first show that the baseline results are at the same level as the original papers. When pre-trained on the MSRAction-3D dataset with our MotionPDE, the fine-tuned backbone shows a consistent improvement over the baseline. The PST-Transformer achieves a 97.3\% accuracy on the MSRAction-3D dataset, surpassing the results of end-to-end training. Additionally, the pre-trained model with MotionPDE also helps to improve the baseline performance on other datasets. For example, on the UTD-MHAD datasets, the PST-Transformer achieves a 4.84\% accuracy gain, proving that the spatial-temporal correlations learned by our MotionPDE are universal and dataset-independent, and therefore easily transferable. 

When pre-trained on the large-scale NTU RGB+D dataset, we observe that the model significantly boosts the performance on the UTD-MHAD dataset but does not adapt as well to the MSRAction-3D dataset, resulting in a slight performance drop. To address the problem, we further test the pre-training on a dense setting of NTU RGB+D by setting the frame sampling stride to 1. The model is pre-trained for 3 epochs with all other settings unchanged. Our findings indicate that the sampling ratio of the dataset, rather than the dataset itself, has a more substantial impact on the pre-training performance. Dense sampling during pre-training proves effective for datasets with sparse data, such as UTD-MHAD, but does not necessarily benefit datasets with dense sampling like MSRAction-3D.

\subsection{Model Efficiency and Parameter Size}
\label{sec:param}

The potential complexity of the MotionPDE model primarily arises from the attention mechanism, which is calculated as $\mathcal{O}(N^2 \cdot D)$ for $N=clip~length \times point~numbers / stride$ and dimension $D$. Other factors contributing to the potential complexity include sinusoidal transformations, concatenations based on input dimensions, and trigonometric functions. Despite these theoretical complexities, the actual resource usage is quite manageable. In practice, the clip length is typically set to 24 and remains constant during backbone inference. The point number is set to 2048 and aggregated to 64 after backbone inference. The hidden dim $D$ for the attention layer is 256, with a backbone dim 2048. 

\begin{table}[htbp]
\caption{The model efficiency and parameters of MotionPDE, using the default settings, are consistent across various backbones. For illustration purposes, we present a comparison with PSTNet as an example.}
\label{tab:params}
\centering
\begin{tabular}{lccc}
\toprule
default setting & parameters(M) & FLOPs(G) & Inference time(ms)\\
\midrule
PSTNet \cite{fan2021pstnet} & 8.44 & 54.09 & 63.88 \\
    {\quad + \textbf{MotionPDE}} & +0.54 & +0.81 & +1.18 \\
\bottomrule   
\end{tabular}
\vspace{-0.3cm}
\end{table}

\begin{figure}[htbp]
\begin{center}
\includegraphics[width=\linewidth]{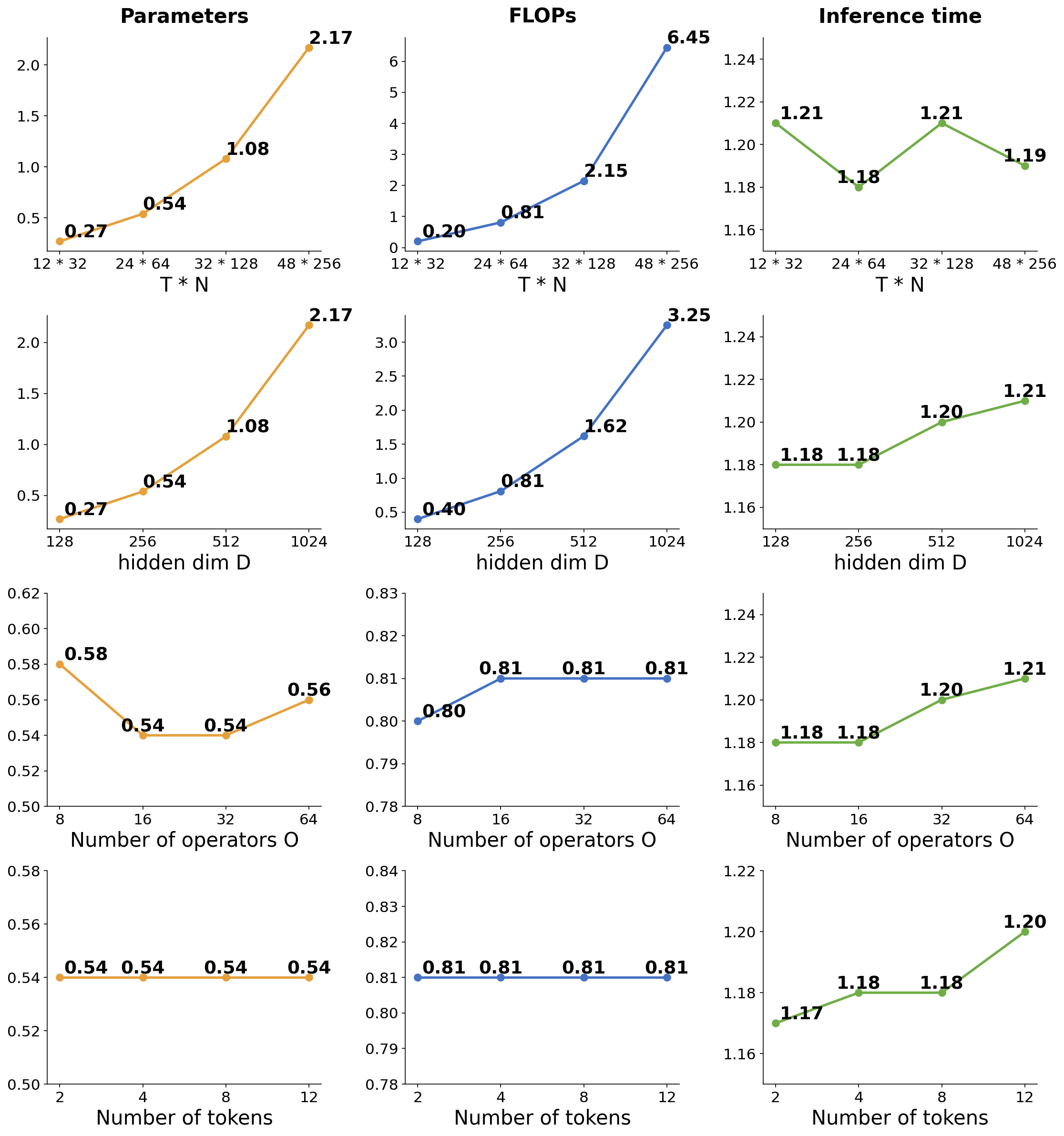}
\end{center}
\vspace{-0.3cm}
\caption{Model efficiency with scale-up setting. }
\label{fig:params}
\end{figure}

From Table \ref{tab:params} we can see that the extra parameters, FLOPs, and inference time caused by the MotionPDE are almost negligible. To better prove our efficiency, we further scale up the data frames, sampling points (aggregated), and other settings, including the hidden dimension of the attention layer, the number of operators, and the number of hidden tokens. We change one parameter at a time. Results are shown in Figure \ref{fig:params}. From the last column, we can see that MotionPDE maintains fast inference in all scenarios. When scaling up the data frames and sampling points (T$\times$N), the model parameters and FLOPs increase quadratically as expected, but still at a very lightweight level. The parameters and FLOPs caused by the MotionPDE go up to only 2.17M and 6.45G with the input shape of 48$\times$256 (which represents 48 frames and 8192 original points when using the same sampling stride). When scaling up the hidden dimension, the model parameters and FLOPs also increase linearly as expected. Furthermore, from the last two rows, we found that the configuration of the operator-learning model (from Section \ref{sec:operatorlearnin}) has minimal impact on the model's efficiency, which proves the all-around lightness of our MotionPDE.

\subsection{Ablation study}
\label{sec:ablation}

\subsubsection{Module Setting of MotionPDE}

We apply ablation studies to the setting of our MotionPDE. We show the ablation results in Table \ref{tab:ablation}. First, we omit the spectral method in Equation \ref{eq:output} and use the simple attention model as the solving model to demonstrate the subtlety of our design. We find that directly omitting the spectral method will lead to a significant drop in performance, while still outperforming the baseline model. The spectral method is essential for maintaining the effectiveness and efficiency of our MotionPDE in achieving optimal enhancements. We further test different hyperparameter settings in our MotionPDE. While the model may exhibit some sensitivity to these parameters, it consistently improves upon the baseline model. This consistent performance across different hyperparameter configurations, along with the dataset-independent and task-independent results shown in the previous sections, suggests that our approach is both robust and universally applicable.

\begin{table}[htbp]
\caption{Ablation study of the impact of the spectral method in our MotionPDE. The experiments are conducted on the MSRAction-3D benchmark with the PSTNet backbone with end-to-end training. In the full MotionPDE module setting, we set the hidden dim D to 512, the number of operators O to 16, and the number of tokens to 4.}
\vspace{-0.3cm}
\label{tab:ablation}
\begin{center}
{\begin{tabular}{lccr}
\toprule
\multicolumn{3}{l}{Settings} & Accuracy(\%)  \\
\midrule
\multicolumn{3}{l}{PSTNet \cite{fan2021pstnet}} & 91.2 \\
\multicolumn{3}{l}{\quad + \textbf{full MotionPDE module}} & \textbf{94.28} \\
\midrule
\multirow{10}{*}{PDE-solving core} & \multicolumn{2}{c}{w/o spectral method} & 92.25 \\
\cmidrule{2-3}
& \multirow{3}{*}{hidden dim D} & 128 & 92.76 \\ 
&  & 512 & 93.03 \\
&  & 1024 & 92.92 \\
\cmidrule{2-3}
& \multirow{3}{*}{operators O} & 8 & 92.92 \\ 
&  & 24 & 93.60 \\
&  & 32 & 91.91 \\
\cmidrule{2-3}
& \multirow{3}{*}{tokens} & 2 & 91.58 \\ 
&  & 6 & 93.60 \\
&  & 8 & 92.94 \\
\bottomrule
\end{tabular}}
\end{center}
\vspace{-0.3cm}
\end{table}

\subsubsection{Pooling Operation}

Note that the max pooling operation along the temporal dimension (the $\text{max-pool}_s$ in Figure \ref{fig:motionpde}), implicitly assumes that the $i$-th point across all frames is spatially aligned or corresponds to a similar physical location over time. While this assumption may not strictly hold in general due to the unordered and dynamic nature of point cloud videos, we have the following clarifications and observations:

First of all, we design an ablation study to compare the performance of our model with and without the pooling operation. We keep $F_t \in \mathbb{R}^{T' \times C}$ as its original way and use $F_s = \text{backbone feature} \in \mathbb{R}^{T' \times N' \times C}$ without pooling. This modification does not affect our modeling process because both the $F_t$ and the $F_s$ are projected into the latent space in the subsequent steps. We show the ablation in the following Table \ref{tab:pooling}.

\begin{table}[htbp]
    \centering
    \caption{Ablation study on the max pooling operation along the temporal dimension. Results on the MSRAction-3D dataset \cite{li2010msraction} are reported.}
    \renewcommand{\arraystretch}{1.}
    \begin{tabular}{l|c|c}
    \toprule
        Method & w/ pooling  & w/o pooling\\
    \midrule
        PSTNet \cite{fan2021pstnet} + \textbf{MotionPDE} & 94.28 & 94.95 \\
        PST-Transformer \cite{fan2023psttransformer} + \textbf{MotionPDE} & 96.63 & 95.96 \\
    \bottomrule
    \end{tabular}
    \label{tab:pooling}
\end{table}

The results showed only slight performance fluctuations (in comparison to our improvements achieved over the baseline) when the max pooling was removed, suggesting that this approximation does not critically affect overall performance. On the other hand, the results also indicate that removing this pooling operation does not lead to a significant performance improvement.

Secondly, in our setting, each input video clip spans a relatively short duration, during which the motion typically undergoes limited deformation. The point sets used in the pooling (\( N' \ll N \)) are generated through downsampling and local aggregation, inherently encoding neighborhood-level information rather than relying on precise point-wise tracking. Furthermore, the backbone models usually include global feature aggregation, which helps promote feature consistency across frames, even in the absence of explicit correspondence.

Lastly, the pooling serves a practical role in compressing feature dimensions and extracting stable spatial representations from potentially noisy or unordered inputs. While this strategy does not enforce exact temporal alignment, it offers a computationally efficient and empirically effective means of capturing aggregated spatial features. Our findings support it as a reasonable design trade-off within the architecture. 

\subsubsection{Performance Under Attacks}

\begin{figure}[htbp]
\begin{center}
\includegraphics[width=\linewidth]{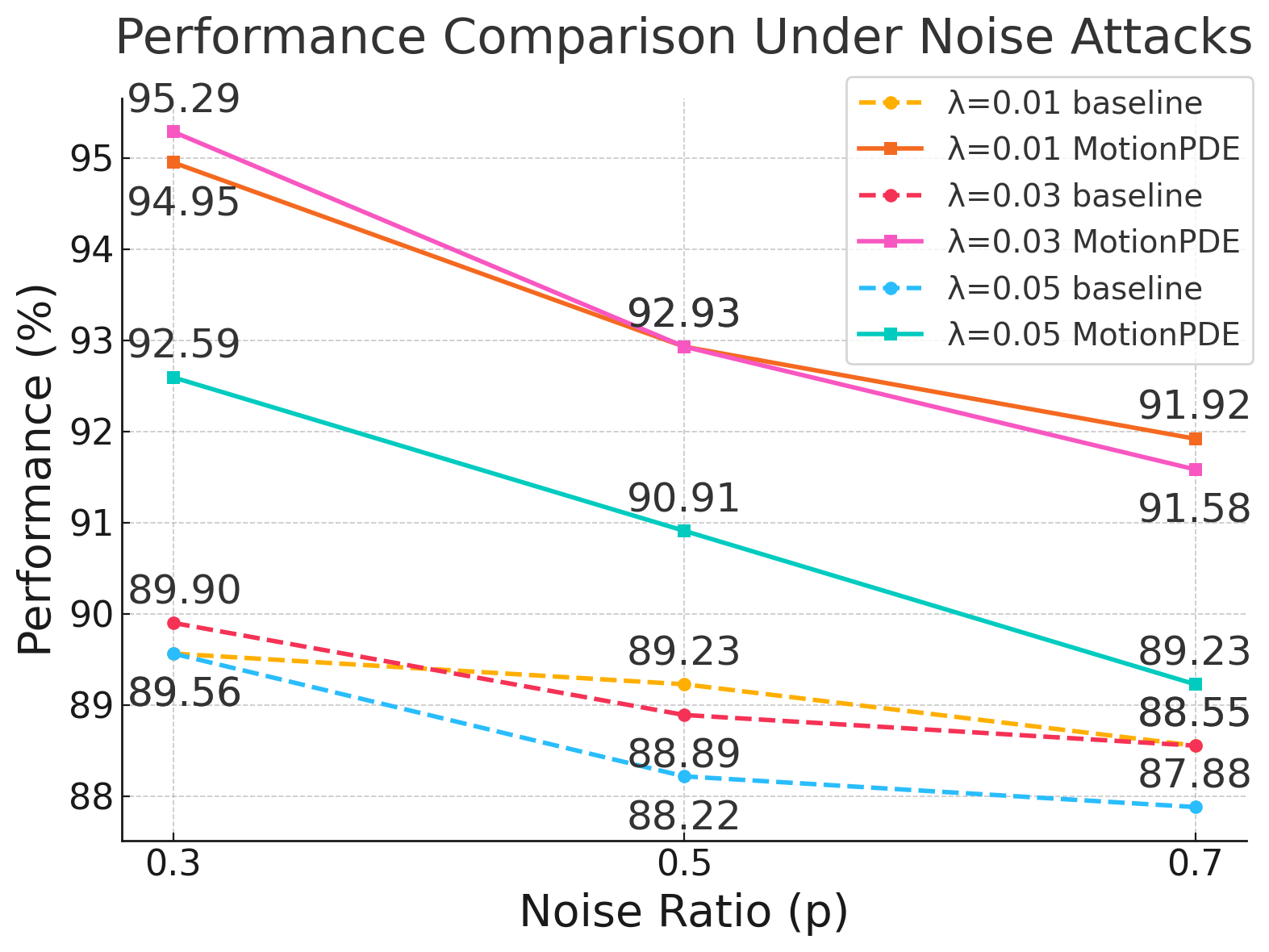}
\end{center}
\vspace{-0.3cm}
\caption{A performance comparison on the MSRAction-3D dataset under noise attacks.}
\label{fig:attackmsr}
\vspace{-0.3cm}
\end{figure}

\begin{figure}[htbp]
\begin{center}
\includegraphics[width=\linewidth]{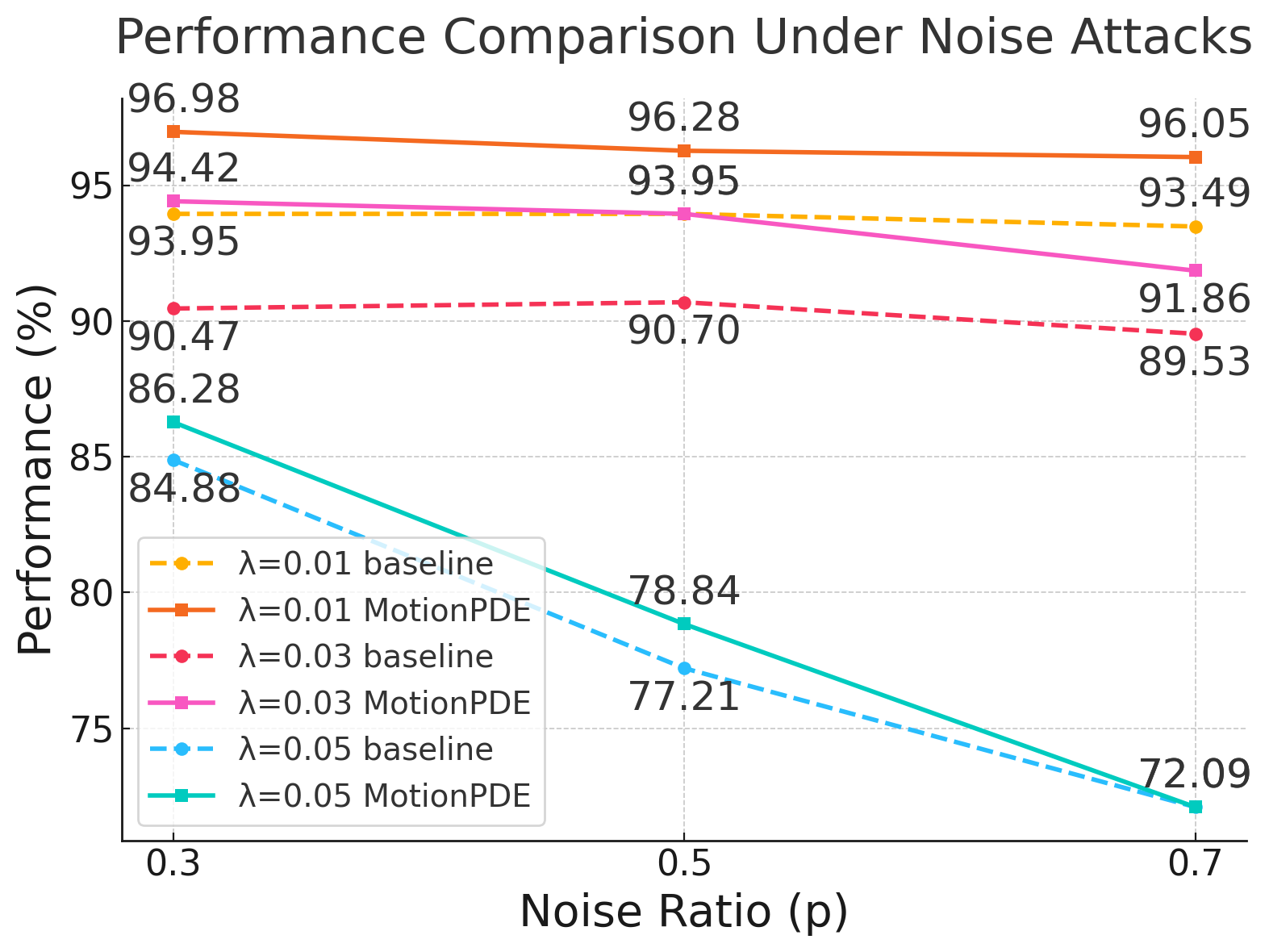}
\end{center}
\vspace{-0.3cm}
\caption{Performance comparison on UTS-MHAD dataset under noise attacks (same legend as Figure \ref{fig:attackmsr} is used).}
\label{fig:attackutd}
\vspace{-0.3cm}
\end{figure}

Point cloud data is known for its sensitivity to noise. To evaluate the robustness of our model, we conducted noise attacks on models trained with the MSRAction-3D and UTD-MHAD datasets by introducing noise to the original points in the testing data. We tested different noise ratios and levels, and the results are presented in Figure \ref{fig:attackmsr} and Figure \ref{fig:attackutd}.

Follows the general principles of prior works \cite{aoki2019pointnetlk,sun2023benchmarking}, we simulate noisy point cloud video data by injecting controlled Gaussian noise into a subset of points. Specifically, we randomly select a fixed proportion \(p \in (0,1)\) of the points in each frame and replace their 3D coordinates with Gaussian noise sampled from \(\mathcal{N}(\mathbf{0}, \lambda^2 \mathbf{I})\), where \(\lambda > 0\) controls the noise intensity and \(\mathbf{I} \in \mathbb{R}^{3 \times 3}\) is the identity matrix. Specifically, for each frame \(t\), a set of indices \(\mathcal{I}_t \subset \{1, \ldots, N\}\) of size \(\lfloor pN \rfloor\) is randomly selected, and the noisy clip \(V'\) is generated as follows:
\begin{gather}
V'_{t, i} =
\begin{cases}
V_{t, i}, & \text{if } i \notin \mathcal{I}_t \\
\boldsymbol{\epsilon}_{t, i}, \quad \boldsymbol{\epsilon}_{t, i} \sim \mathcal{N}(\mathbf{0}, \lambda^2 \mathbf{I}), & \text{if } i \in \mathcal{I}_t
\end{cases}
\end{gather}


\begin{figure*}[htbp]
\begin{center}
\includegraphics[width=\linewidth]{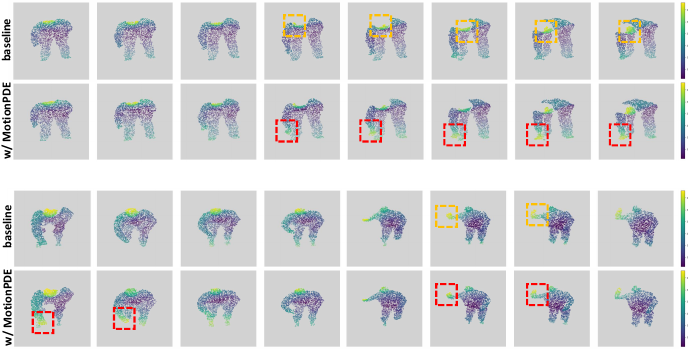}
\end{center}
\vspace{-0.3cm}
\caption{Visualization on different clip samples of "Pickup \& Throw". Clips are sampled from the same video data from the test set of the MSRAction-3D dataset. Brighter colors indicate stronger feature response.}
\label{fig:featvis}
\end{figure*}

Our model trained with MotionPDE exhibits sustained effectiveness under noise attacks in both datasets. With low-level noise (e.g.,  $p = 0.3$ at a noise level $\lambda = 0.01$), the model performs even better than reported. These results demonstrate that MotionPDE provides a significant improvement over the baseline models under various noise conditions. At higher noise levels (e.g., noise levels $\lambda = 0.03$ and $\lambda = 0.05$), MotionPDE still maintains higher performance, compared to the baseline, indicating its effectiveness in preserving critical spatial-temporal correlations in noisy environments.

\subsection{Analysis and Visualization}
\label{sec:analysis}

\subsubsection{Feature Visualization} 

To further prove that our MotionPDE can bring extra regularisation to spatial-temporal correlation learning, we visualize the learned features from the networks to investigate the impact of MotionPDE. We experiment on the MSRAction-3D using PSTNet \cite{fan2021pstnet} as the baseline. Results are shown in Figure \ref{fig:featvis}. Since ReLU is adopted in the PSTNet backbone, larger values with brighter colors indicate stronger feature response.

As discussed in both the PSTNet \cite{fan2021pstnet} and PST-Transformer \cite{fan2023psttransformer}, these models are designed to capture point dynamics, yielding higher activation in regions with motion. Our MotionPDE further enhances this capability, leading to more pronounced feature responses in dynamic areas.

\begin{table}[htbp]
\caption{Clip accuracy compared to the baseline. The results are reported using our own implementation.}
\label{tab:clipacc}
\centering
\begin{tabular}{lcc}
\toprule
Methods & MSR & UTD \\
\midrule
PSTNet \cite{fan2021pstnet} & 85.06 & 90.95 \\
{\quad + \textbf{MotionPDE}} & 88.39 & 92.57 \\
\bottomrule   
\end{tabular}
\end{table}

\begin{figure*}[htbp]
\begin{center}
\includegraphics[width=\linewidth]{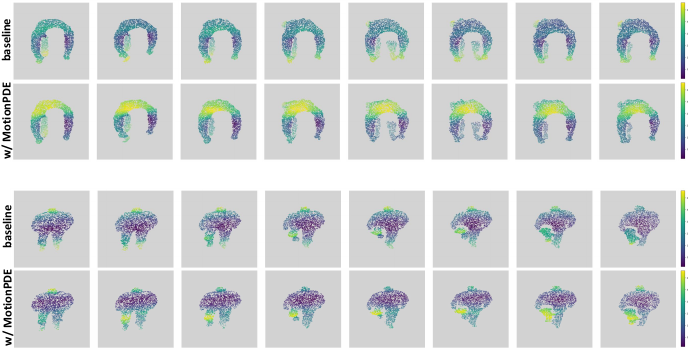}
\end{center}
\vspace{-0.3cm}
\caption{Visualization of "Bend" and "Forward kick" action clips. The clips are sampled from the test set of the MSRAction-3D dataset. The brighter colors indicate higher feature response.}
\label{fig:featvis1}
\end{figure*}

\begin{figure*}[htbp]
\begin{center}
\includegraphics[width=\linewidth]{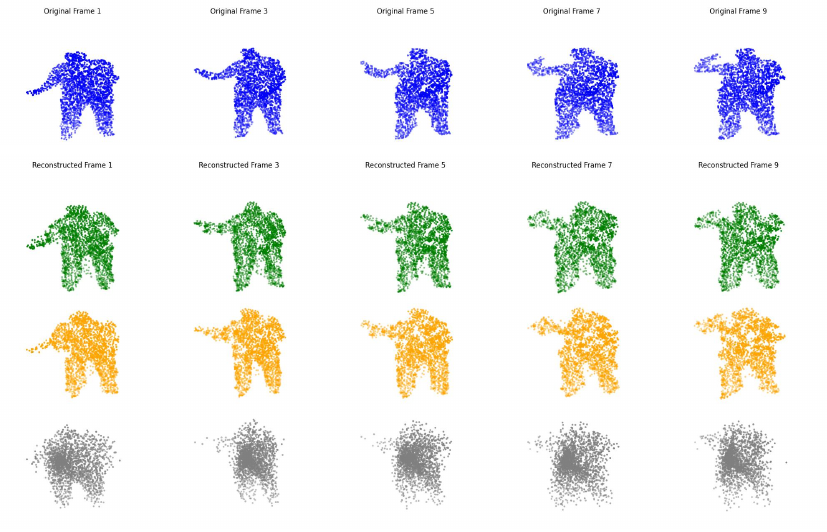}
\end{center}
\vspace{-0.5cm}
\caption{Visualization of the reconstruction results. The first row is the \textcolor{blue}{ground truth}, followed by the reconstruction result of the \textcolor[rgb]{0., 0.5, 0.}{pre-trained model with MotionPDE}, the \textcolor[rgb]{1., 0.65, 0.}{end-to-end trained model with MotionPDE}, and the \textcolor{gray}{baseline} \cite{fan2021pstnet}.}
\label{fig:reconstruction}
\end{figure*}

The model trained with our MotionPDE gains a more comprehensive understanding of complex movements, effectively addressing the challenges posed by data incompleteness due to clip sampling. For instance, when analyzing an action like ``Pickup \& Throw", the baseline model inaccurately focuses on the human head (highlighted by the \textcolor{orange}{orange box}) in the first row of Figure \ref{fig:featvis}, whereas our model accurately tracks the human hand (highlighted by the \textcolor{red}{red box} in the second row). This discrepancy arises because the incomplete data clips mislead the baseline model. The baseline model only highlights clips related to the `throw' action in a mixed data clip (\textcolor{orange}{orange box} in the third row), whereas our model correctly comprehends the entire sequence from `picking' to `throwing' (\textcolor{red}{red box} in the last row). This demonstrates the superior ability of our MotionPDE-enhanced model to capture the full range of motion in point cloud videos, even with incomplete data. Such improvement will lead to a higher clip accuracy (result shown in Table \ref{tab:clipacc}) and a significant reduction in misclassification compared to the baseline (also proved by the confusion matrix provided in Section \ref{sec:ConfusionMatrix}). Overall, these results underscore the utility of incorporating MotionPDE into models for more accurate and reliable point cloud video analysis.

Additionally, such a comprehensive understanding also brings more nuanced enhancements compared to the baseline. Seeing the top section of Figure \ref{fig:featvis1}, our model highlights the body itself rather than the drooping hand in the ``Bend" action. In the bottom section, concerning the ``Forward kick" action, our model's attention shifts to the leg region earlier and also responds stronger (brighter) than the baseline. Such focus suggests that our model has a better grasp of the essential aspects of the action, rather than just tracking the moving parts indiscriminately. 

Overall, these observations underline that our model, enhanced with MotionPDE, captures the fundamental nature of the actions being performed, rather than merely responding to moving segments. This attention to detail is crucial, as it enables the model to understand the underlying meaning of the actions rather than just detecting motion. 

\begin{table*}[htbp]
\caption{Additional action recognition results on unsuitable networks. Our leads to a performance drop in most cases due to the absence of localized spatial information.}
\label{tab:addresult}
\begin{center}
\renewcommand{\arraystretch}{1.}
\resizebox{0.9\linewidth}{!}{
\begin{tabular}{lcccccc}
\toprule
    \multirow{2}{*}{Methods} & \multirow{2}{*}{MSR} & \multicolumn{2}{c}{NTU60} & \multicolumn{2}{c}{NTU120} & \multirow{2}{*}{UTD} \\
    & & Cross-Subject & Cross-View & Cross-Subject & Cross-Setup\\
\midrule
     SequenctialPointNet \cite{li2021sequentialpointnet} & 91.94 & 90.3 & 97.6 & 83.5 & 95.4 & 92.31 \\
    {\quad + \textbf{MotionPDE}} & 89.22 & 88.4 & 96.1 & 83.1 & 93.5 & 92.79 \\
    P4Transformer \cite{fan21p4transformer} & 90.94 & 90.2 & 96.4 & 86.4 & 93.5 & - \\
    {\quad + \textbf{MotionPDE}} & 92.25 & 88.6 & 95.8 & 84.4 & 93.0 & - \\
\bottomrule
\end{tabular}
}
\end{center}
\vspace{-0.3cm}
\end{table*}

\subsubsection{Reconstruction Visualization}
\label{sec:reconstructionresult}

As discussed in the introduction section, we also perform reconstruction experiments with our model and compare the visualization results to the baseline \cite{fan2021pstnet}. We freeze the backbone from the trained model and train a simple MLP built with 3 transposed convolution layers as the reconstruction decoder. We then reshape the decoder output to the original point cloud shape and supervise the training process with Chamfer Distance.

Firstly, we compare the reconstruction results presented in Figure \ref{fig:reconstruction}. The figure displays the results from top to bottom as follows: the ground truth, the pre-trained model with MotionPDE (as described in Section \ref{sec:pretrain}), the end-to-end trained model with MotionPDE (detailed in Section \ref{sec:msraction3d}), and the baseline. From the second row, we find that the reconstruction results of the pre-trained model with MotionPDE are very close to the ground truth. We demonstrate that using only our contrastive loss, MotionPDE can guide the backbone to learn the spatio-temporal correlations in point cloud videos, resulting in an impressive reconstruction performance. This comparative analysis highlights the superior reconstruction capability of models utilizing MotionPDE, illustrating its effectiveness in learning spatial-temporal correlations. The spatio-temporal correlation learning is task-independent and persists in the end-to-end trained model with MotionPDE. Compared to the reconstruction results of the baseline model, the superiority of our MotionPDE is bolstered.

\subsubsection{Confusion Matrix}
\label{sec:ConfusionMatrix}

\begin{figure}[htbp]
    \begin{center}
    \subfloat{\includegraphics[width=\linewidth]{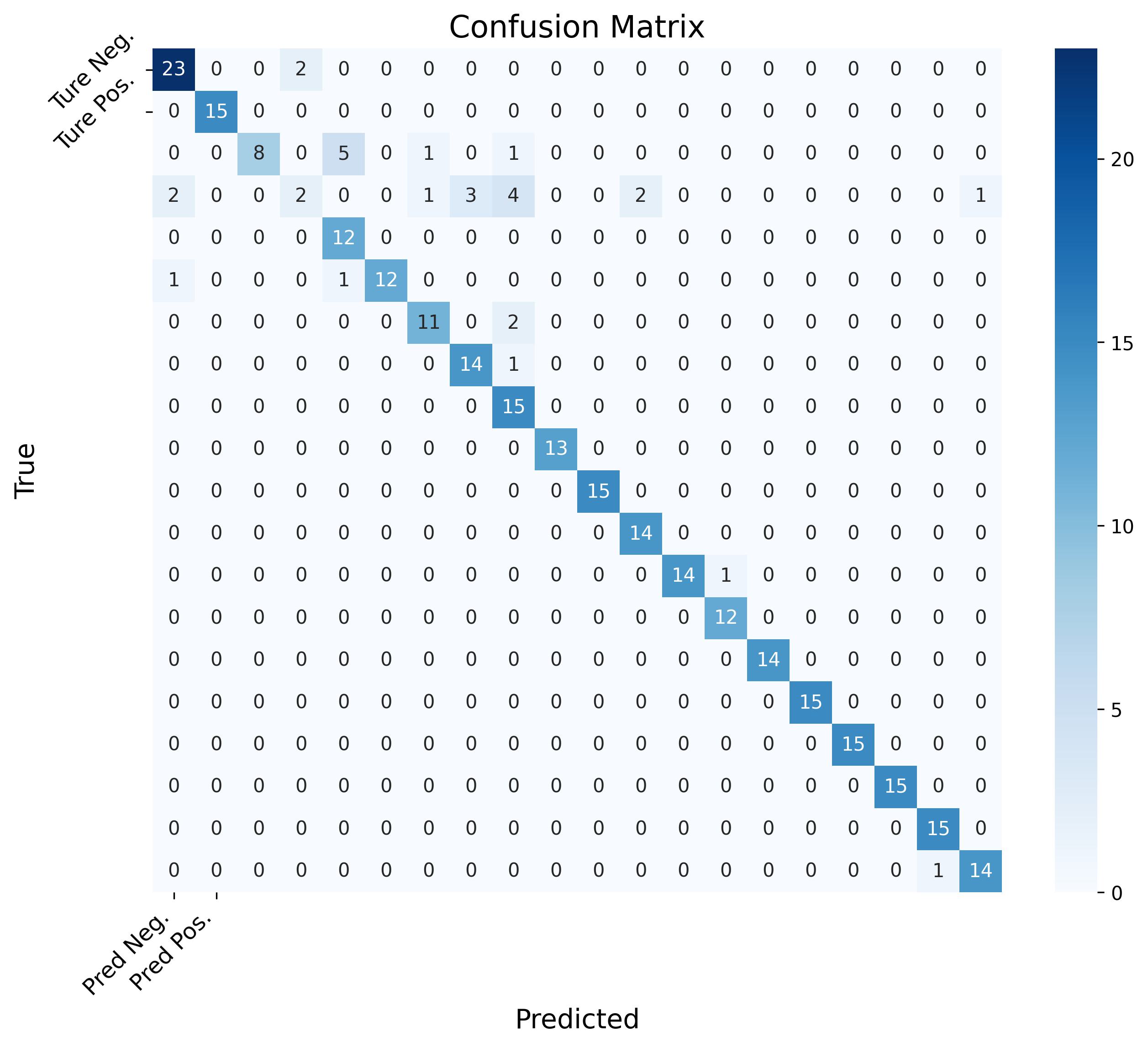}}\\
    \subfloat{\includegraphics[width=\linewidth]{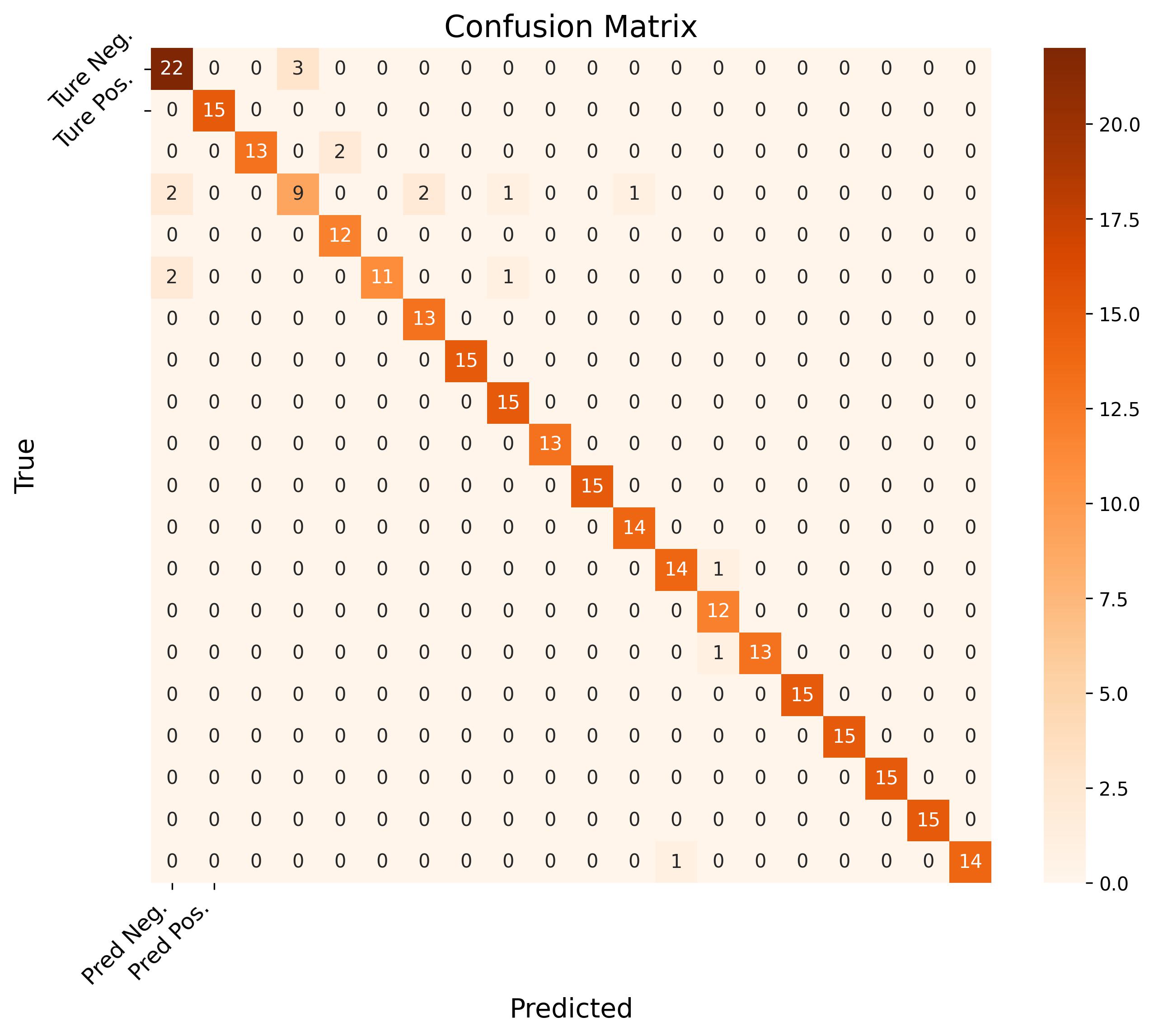}}
    \caption{Confusion matrix computed on MSRAction-3D. Upper: PSTNet baseline. Bottom: End-to-end trained model with MotionPDE. }
    \label{fig:cmmsr}
    \end{center}
\vspace{-0.5cm}
\end{figure}

To further substantiate the effectiveness of our approach, we present additional evidence through the analysis of the Confusion Matrix. The Confusion Matrix provides a detailed breakdown of the classification performance, illustrating the model's accuracy in distinguishing between different classes. The analysis in Figure \ref{fig:cmmsr} offers a quantitative measure of the improvements achieved by incorporating MotionPDE.

The Confusion Matrix reveals a significant reduction in misclassifications compared to the baseline, mostly because of the comprehensive understanding of the point cloud video (details in the feature visualization section).

\subsubsection{Discussion on limitations}

To discuss our limitation, we adopt our MotionPDE to the SequentialPointNet \cite{li2021sequentialpointnet} and the P4Transformer \cite{fan21p4transformer}, which share a similar structure that compresses the frame information into a global vector. Results are shown in Table \ref{tab:addresult}. While these two networks are structurally compatible with the integration of MotionPDE, they compress each frame into a global feature vector (the shape of the backbone feature is $T' \times 1 \times C$), eliminating the local spatial structure. Since MotionPDE relies on modeling the spatio-temporal variations, the lack of local spatial granularity makes integration infeasible. Specifically, our MotionPDE achieved performance improvements when integrated with P4Transformer on the MSRAction-3D \cite{li2010msraction} dataset and with SequentialPointNet on the UTD-MHAD \cite{utddataset} dataset, but exhibited performance degradation in other experiments. Adopting our MotionPDE in unsuitable networks leads to a performance drop in most cases due to the absence of localized spatial information. And therefore, the PDE model leads to the wrong regularization direction.

Aside from that, MotionPDE exhibits excellent performance across different backbone models and a wide spectrum of tasks (see from Table \ref{tab:actrecognition} to Table \ref{tab:selfsupervised}), further confirming its strong research contribution. Our work is the first to introduce a PDE-based model into point-cloud video analysis and to adapt it across the majority of standard baseline architectures. In future work, we will investigate deeper integration with more complex frameworks to further validate the versatility and adaptability of MotionPDE.

\section{Conclusion}

This paper presented MotionPDE, a novel enhancement approach for point cloud video representation learning through PDE-solving. Our work is the first to apply PDE modeling to point cloud video data, drawing a compelling parallel between fluid dynamics and the understanding of point cloud videos. By regularizing spatial-temporal correlation learning within the backbone model, MotionPDE significantly boosts the performance of point cloud video learning, achieving state-of-the-art results with minimal additional computational cost and parameter overhead. Extensive experiments across multiple benchmarks validated the effectiveness and robustness of our approach, highlighting its ability to elevate existing methods to new levels of accuracy. The impressive outcomes of MotionPDE emphasized the potential of PDE-based methodologies in tackling complex spatial-temporal data challenges, opening new directions for innovation in this domain. 
In particular, we see strong potential for extending MotionPDE to more complex point cloud video tasks such as 4D semantic segmentation and scene flow estimation. These tasks involve modeling dense, scene-level dynamics that may require more expressive PDE formulations. We consider this a promising and challenging direction for future exploration.



405\bibliography{trans.bib}
\bibliographystyle{IEEEtran}


\begin{IEEEbiography}[{\includegraphics[width=1in,height=1.25in,clip,keepaspectratio]{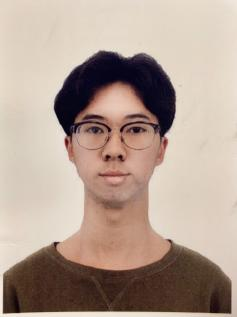}}]{Zhuoxu Huang} is a senior Ph.D. student in Computer Science at Aberystwyth University, UK. He received a bachelor’s degree from Wuhan University, China. He is a Vision Graphics and Visualisation Group member at Aberystwyth University. His research interests include 3D vision, video analysis, and multimodal LLM. 
\end{IEEEbiography}

\begin{IEEEbiography}[{\includegraphics[width=1in,height=1.25in,clip,keepaspectratio]{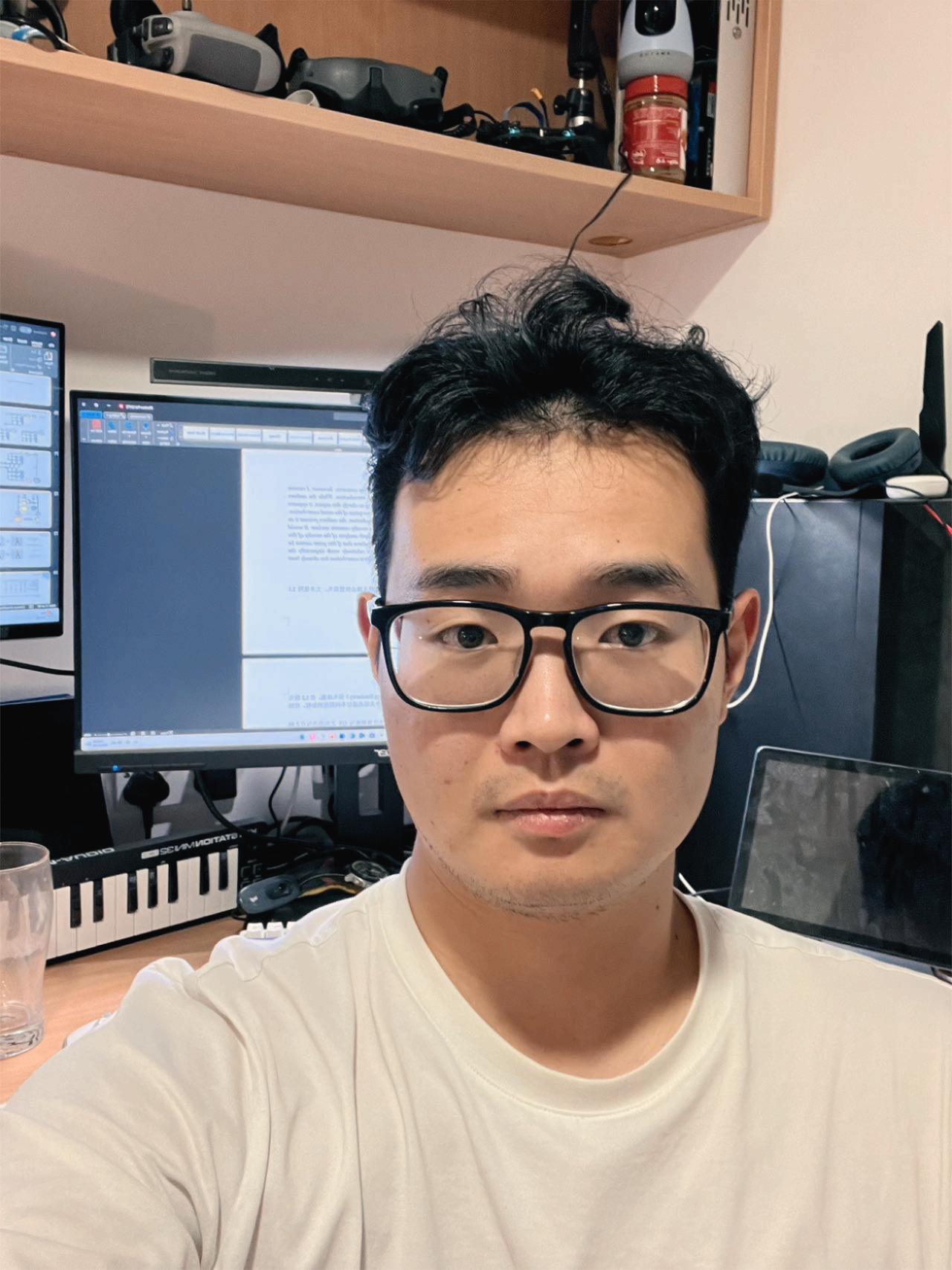}}]{Zhenkun Fan}, currently pursuing a Ph.D. degree in Computer Science at Aberystwyth University, UK. His research focuses on areas of computer vision such as pose estimation and semantic segmentation, He received his master's degree in computer science from Ocean University of China in 2022 and a B.E. degree in Computer Science from Shandong University of Science and Technology in 2019.
\end{IEEEbiography}

\begin{IEEEbiography}[{\includegraphics[width=1in,height=1.25in,clip,keepaspectratio]{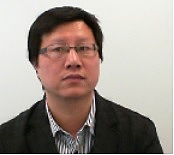}}]{Jungong Han} is a professor in the Department of Automation at Tsinghua University. He also holds an Honorary Professorship at Aberystwyth University, and the University of Warwick, UK. His research interests include computer vision, artificial intelligence, and machine learning. He is a Fellow of IAPR and a Fellow of AAIA. He serves as the Associate Editor for many prestigious journals, such as IEEE Transactions on Multimedia, IEEE Transactions on Image Processing, IEEE Transactions on Neural Networks and Learning Systems, IEEE Transactions on Circuits and Systems for Video Technology, and Pattern Recognition.
\end{IEEEbiography}

\begin{IEEEbiography}[{\includegraphics[width=1in,height=1.25in,clip,keepaspectratio]{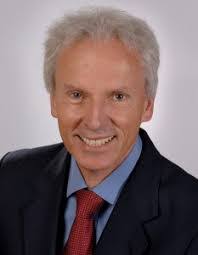}}]{Josef Kittler} (Life Member, IEEE) received the B.A., Ph.D., and D.Sc. degrees from the University of Cambridge in 1971, 1974, and 1991, respectively. He is currently a distinguished professor of machine intelligence at the Centre for Vision, Speech and Signal Processing, University of Surrey, Guildford, U.K. His research interests include biometrics, video, image database retrieval, medical image analysis, and cognitive vision.
\end{IEEEbiography}



\flushend

\end{document}